\documentclass[11pt,final]{article} 

\usepackage{amsfonts}
\usepackage{amssymb}
\usepackage{amsmath}
\usepackage{graphics}
\usepackage{float}
\usepackage{graphicx}
\usepackage{algorithm2e}
\usepackage{hyperref}

\hypersetup{
    pdffitwindow=false,     
    pdfstartview={FitH},    
    pdfnewwindow=true,      
    colorlinks=true,       
    linkcolor=red,          
    citecolor=green,        
    filecolor=magenta,      
    urlcolor=cyan           
}

\renewcommand{\baselinestretch}{1.5} 

\def\sep{\unskip; }                  

\def\la#1{\label{eq:#1}}
\def\re#1{(\ref{eq:#1})}
\def\cite#1{\ref{#1}}

\begin{document}

\title{A Comparison of Bayesian Prediction Techniques for Mobile Robot Trajectory Tracking \thanks{\protect \parbox[t]{.95\textwidth}{This work was supported by Conycit of Chile under Fondecyt Grant 11060251.\protect \\[-1mm]
\protect \scriptsize Dept. of Electrical Engineering, Pontificia Universidad Cat\'olica de Chile,\protect \\[-1mm] 
Vicu\~na Mackenna 4860, Casilla 306-22, Santiago, Chile\protect \\[-1mm] 
 E-mail: \texttt{\{mtorrest\}@ing.puc.cl} }}}

\author{J.L. Peralta-Cabezas, \hspace{1ex} M. Torres-Torriti and\hspace{1ex} M. Guarini-Hermann}

\date{}

\maketitle

\begin{abstract}
This paper presents a performance comparison of different estimation and prediction techniques applied to the problem of tracking multiple robots.  The main performance criteria are the magnitude of the estimation or prediction error, the computational effort and the robustness of each method to non-Gaussian noise.  Among the different techniques compared are the well known Kalman filters and their different variants (e.g. extended and unscented), and the more recent techniques relying on Sequential Monte Carlo Sampling methods, such as particle filters and Gaussian Mixture Sigma Point Particle Filter.\\[3mm]
\end{abstract}

\thispagestyle{empty} 
\vspace{-15mm}        
\parbox[t]{0.99\textwidth}{ 
\vspace{95mm} {\scriptsize \bf Accepted in Robotica (Dec. 2007),
vol. 26, n. 5, pp. 571--585 \copyright\ 2008 Cambridge University Press.}
\bf Cite DOI: \url{https://doi.org/10.1017/S0263574708004153}
\vspace{-130mm}
}


\section{Introduction}\label{sec:introduction}
Filtering and prediction methods to accurately track multiple moving objects under noisy measurements, disturbances and model uncertainty have received much attention during the last decades in a wide variety of fields, such as aeronautics, defense, computer vision, robotics and economics, to name a few$^{\cite{THR05}-\cite{BAR98}}$.  It is a well-known fact that the ability of any control or decision system to meet its targets directly depends on accuracy of the measurements and the state estimation$^{\cite{ASK00}}$.  On the other hand, when the measurements of the controlled variables are available with some delay, the ability to make accurate predictions of the measurements becomes particularly important in order to compensate the measurements' latency$^{\cite{MER04A}-\cite{BOU01}}$. 

In view of the large number of existing approaches (see references in table~\ref{tab:filt}) and the key role of filtering and estimation techniques for accurate tracking, it is important to identify those techniques that have the best performance, not only in terms of accuracy, but also in terms of their ability to handle non-Gaussian noise and their computational cost.  The aim of the present work is thus to evaluate the performance of classic filtering techniques for trajectory prediction and perception-latency compensation, such as the Kalman Filter (KF)$^{\cite{KAL60}}$ and the Extended Kalman Filter (EKF)$^{\cite{JAZ70}}$, as well as more recent ones, such as the Unscented Kalman Filter (UKF)$^{\cite{JUL97}}$, Particle Filtering (PF)$^{\cite{RIS04}}$ and other variants in the class of the so-called Sigma-Point Kalman Filters (SPKF)$^{\cite{MER04A},\cite{ITO00},\cite{NOR00}}$.  The prediction strategies compared in this paper, which are summarized in table~\ref{tab:filt}, can be grouped into two main families: (i) the Kalman Filter based approaches$^{\cite{KAL60},\cite{JAZ70},\cite{JUL97},\cite{NOR00},\cite{ITO00},\cite{MER04A},\cite{MER01}}$ and (ii) the sequential sampling or Particle Filter based approaches$^{\cite{RIS04},\cite{DOU00},\cite{MER00},\cite{MER03}}$.  

{\renewcommand{\baselinestretch}{1.2}
\begin{table}[htbp]
\begin{center}
\small\begin{tabular}{lcccc}
\hline
\parbox[t]{0.25\textwidth}{\centering\textit{Filter}} & 
\textit{Abrev.} & 
\parbox[t]{0.11\textwidth}{\centering\textit{Noise Assumption}} & 
\parbox[t]{0.32\textwidth}{\textit{Linearization Technique}} & 
\textit{Reference} \\[2ex]
\hline
\parbox[t]{0.25\textwidth}{Extended Kalman Filter} & 
EKF & 
Gaussian & 
\parbox[t]{0.32\textwidth}{{\small 1$^{st}$ Order Taylor}} & 
{\small
 ${\cite{KAL60},\cite{JAZ70}}$}\\[2ex]
\parbox[t]{0.25\textwidth}{Unscented Kalman Filter} & 
UKF & 
Gaussian & 
\parbox[t]{0.32\textwidth}{{\small None}} & 
{\small 
 ${\cite{JUL97}}$}\\[2ex] 
\parbox[t]{0.25\textwidth}{\raggedright Central Difference\\[-.3\baselineskip] 
                                        Kalman Filter} & 
CDKF & 
Gaussian & 
\parbox[t]{0.32\textwidth}{{\small 2$^{nd}$ Order Stirling}} & 
{\small 
 ${\cite{ITO00}}$} \\[2ex]
\parbox[t]{0.25\textwidth}{\raggedright Divided Difference\\[-.3\baselineskip] 
                                         1$^{st}$ Order Kalman Filter} & 
DD1 & 
Gaussian & 
\parbox[t]{0.32\textwidth}{\small 1$^{st}$ Order Stirling} & 
{\small ${\cite{NOR00}}$} \\[2ex]
\parbox[t]{0.25\textwidth}{\raggedright Divided Difference\\[-.3\baselineskip]
                                        2$^{nd}$ Order Kalman Filter} & 
DD2 & 
Gaussian & 
\parbox[t]{0.32\textwidth}{{\small 2$^{nd}$ Order Stirling}} & 
{\small ${\cite{NOR00}}$} \\[2ex] 
\parbox[t]{0.25\textwidth}{\raggedright Square Root Unscented\\[-.3\baselineskip]
                                        Kalman Filter} & 
SRUKF & 
Gaussian & 
\parbox[t]{0.32\textwidth}{{\small None}} & 
{\small ${\cite{MER04A},\cite{MER01}}$} \\[2ex]
\parbox[t]{0.25\textwidth}{\raggedright Square Root Central\\[-.3\baselineskip]
                                        Difference Kalman Filter} & 
SRCDKF & 
Gaussian & 
\parbox[t]{0.32\textwidth}{{\small 2$^{nd}$ Order Stirling}} & 
{\small ${\cite{MER04A},\cite{MER01}}$} \\[2ex]
\parbox[t]{0.25\textwidth}{\raggedright Generic Particle Filter} & 
PF & 
none & 
\parbox[t]{0.32\textwidth}{{\small None}} & 
{\small ${\cite{RIS04},\cite{DOU00}}$} \\[2ex]
\parbox[t]{0.25\textwidth}{\raggedright Sigma-Point\\[-.3\baselineskip]
                                        Particle Filter} & 
SPPF & 
Gaussian & 
\parbox[t]{0.32\textwidth}{{\small None (w/SRUKF),\\[-.3\baselineskip]
                                   2$^{nd}$ Order Stirling (w/SRCDKF)}} &
{\small ${\cite{MER00}}$} \\[2ex]
\parbox[t]{0.25\textwidth}{\raggedright Gaussian Mixture Sigma\\[-.3\baselineskip]
                                        Point Particle Filter} & 
GMSPPF & 
Gaussian & 
\parbox[t]{0.32\textwidth}{{\small None (w/SRUKF),\\[-.3\baselineskip]
                                   2$^{nd}$ Order Stirling (w/SRCDKF)}} & 
{\small ${\cite{MER03},\cite{KOT01}}$} \\
\hline
\end{tabular}
\end{center}
\caption{Summary of Bayesian prediction techniques compared in the present work.}
\label{tab:filt}
\end{table}
} 

Although there exist many studies and comparisons in the literature$^{\cite{SSE04},\cite{MER04A},\cite{ITO00},\cite{CUI05},\cite{WRI03},\cite{YUE02}}$, most of these compare just a few approaches and in general do not consider directly issues related to the robustness of the methods to non-Gaussian noise.  The Gaussianity of the disturbances is a common assumption that often simplifies analytic developments, but is far from the actual characteristis of many real systems.  In fact, some of the existing algorithms yield poor estimates or simply diverge when the Gaussianity assumption is not satisfied, while other require a significantly large computation time that may render them unpractical for real-time control and decision applications$^{\cite{MER04A}}$.  The KF and EKF are unarguably the most popular prediction approaches$^{\cite{RIS04}}$.  However, both rely on a first order Taylor series approximation of the nonlinear state space model of the system. Furthermore, both assume that process and measurement noises are identically and independently Gaussian distributed.  Because of these assumptions, the KF or EKF perform poorly when the system dynamics is highly nonlinear and the system noises are non-Gaussian.  These limitations have motivated the development of the  Sigma-Point Kalman Filters (SPKF)$^{\cite{MER04A}}$ and the rebirth of the computationally intensive Sequential Monte Carlo Sampling methods$^{\cite{DOU00}}$ aided by the availability of more powerful computers.

The main contribution of this paper is the evaluation and comparison of the performance of the ten different filtering strategies listed in table~\ref{tab:filt} as applied to a non-linear system with measurements subject to non-Gaussian noise.  The performance of each filter is measured in terms of the root mean squared-error of the filter's prediction error, the filter's effectiveness to handle non-Gaussian noise and its computational cost.  Concisely, but in a precise manner, the paper also presents the general algorithm of the KF-based and the PF-based filtering approaches.  The system employed as test bench is an ensemble of fast moving RoboCup F-180 robots$^{\cite{GLO03}}$.  The performance of the filters is first evaluated through numerical simulations of the robots' trajectory prediction.  The best performing approach is then tested with the real robots.  In this context, the performance is discussed in terms of the ability of the estimation strategies to filter perception errors and compensate perception delays in real-time.  The results are valuable not only to applications related to robotic soccer competitions, but to a wide range of fields characterized by non-linear dynamics/non-Gaussian disturbances that require accurate state trajectory predictions and the compensation of perception and computation delays.  Examples of these fields include the process control industry$^{\cite{ASK00}}$, the visual tracking applications$^{\cite{CUE05A},\cite{CUE05B}}$ and the autonomous localization and navigation$^{\cite{MER04A},\cite{BOU01},\cite{DON07}}$.

The paper is organized as follows.  Section 2 presents the mathematical background on which the KF-based and PF-based filtering approaches rely.  The derivation of the kinematic and dynamic equations of a three-wheel omni-directional robot employed in the filters implementation is presented in section 3.  The testing methodology and results are presented in sections 4 and 5, respectively.  The conclusions of this work and a discussion of some aspects concerning ongoing research are presented in section 6.

\section{Bayesian Estimation}\label{sec:bayesian_estimation}
In order to define the problem of Bayesian estimation, let $x_k\in\mathbb{R}^{n}$ denote the $n$-dimensional vector representing the state of some system at time $t_k=kT$, $k\in\mathbb{N}$, where $T>0$ is the sampling period (assumed constant for simplicity of exposition), and consider the evolution of the state according to a discrete-time stochastic {\em model equation}:
\begin{eqnarray}
x_k=f_{k-1}(x_{k-1},u_{k-1},v_{k-1})\la{state_space_model}
\end{eqnarray}
where $f_{k-1}$ is a known function of the state $x_{k-1}$, an $m$-dimensional input vector $u_{k-1}\in\mathbb{R}^m$, whose purpose is to drive the state of the system, and an $n$-dimensional process noise $v_{k-1}\in\mathbb{R}^n$ that accounts for external disturbances and mismodelling effects.  The function $f_{k-1}$ is often nonlinear and also may  be time-dependent.

The optimal estimation problem consists in finding the best estimate of the state $x_k$ of the stochastic process~\re{state_space_model} whose actual state cannot be directly observed, but can be inferred from the measurements of the system outputs $y_k\in\mathbb{R}^p$.  The later are related to the system's state by the so-called {\em measurement equation}, which is also known as {\em sensor or observation model}:
\begin{eqnarray}
y_k=h_k(x_{k-l},w_{k-l})\la{measurement_model}
\end{eqnarray}
where $h_k$ is a known, possibly nonlinear and time-dependent, function of the state $x_{k-l}$, at some time $t_{k-l}=(k-l)T$, and a $p$-dimensional measurement noise $w_{k-l}\in\mathbb{R}^n$ vector that accounts for sensor noise and measurement uncertainties.  Here $l\in\mathbb{N}$ represents a possibly non-zero delay introduced by limitations in the speed at which measurements can be acquired.

In the Bayesian formulation, the optimal state estimate at time $t_k$, often denoted by $\hat{x}_k$, is the one that maximizes the {\em a posteriori} conditional probability of the state having certain value given the system output measurements.  More rigorously, if $Y_{k}\stackrel{def}{=}\{y_i,i=1,2,\ldots,k\}$ denotes the set of all sensor measurements up to time $t_k$, then the optimal state estimate in the Bayesian framework is the one that maximizes $p(x_{k}|Y_{k})$, or equivalently, that which minimizes the expected error between the actual state $x_k$ and the estimate $\hat{x}_k$.  Such optimal estimate is uniquely specified by the {\em conditional mean} or {\em minimum variance} estimate (see Thm. 3.1, in reference~{\cite{AND05}) given by:
\begin{eqnarray}
\hat{x}_{k}\stackrel{def}{=}E\left[ x_{k}|Y_{k}\right] =\int x_{k}p\left(x_{k}|Y_{k}\right) dx_{k}\la{EXP}
\end{eqnarray}

Central to the calculation of the optimal estimate is the calculation of the posterior probability density $p(x_k|Y_k)$, which by Bayes' rule can be computed recursively as new measurements arrive according to:
\begin{eqnarray}
p\left( x_{k}|Y_{k}\right) =C\cdot p\left( y_{k}|x_{k}\right) p\left(
x_{k}|Y_{k-1}\right)\la{PROB1}
\end{eqnarray}
where
\[p\left( x_{k} |Y_{k-1} \right) =\int p\left( x_{k} |x_{k-1} \right) p\left(
x_{k-1} |Y_{k-1} \right) dx_{k-1}\]
and
\[C=\left( \int p\left( y_{k}|x_{k}\right) p\left( x_{k}|Y_{k-1}\right)
dx_{k}\right) ^{-1}.\]

If the process is governed by a linear stochastic difference (or differential) equation with Gaussian noises, then the optimal estimate~\re{EXP} can be computed recursively using the well known Kalman Filter$^{\cite{KAL60},\cite{AND05}}$.  However, most real world systems are not linear and are subject to non-Gaussian noises. Thus analytical closed form expressions for the estimate~\re{EXP} usually do not exist, not even in recursive form, and only suboptimal solutions may be computed to approximate in some sense the optimal solution.  Among these approximated solutions are the Extended Kalman Filter (EKF)$^{\cite{RIS04},\cite{JAZ70}}$, Gaussian Sum Filters (GSF)$^{\cite{MER03},\cite{KOT01}}$, Sigma-Point Kalman Filters (SPKF)$^{\cite{MER04A}}$, and Sequential Monte Carlo (SMC) methods that give origin to the Particle Filter (PF) and its variants$^{\cite{RIS04},\cite{MER00}-\cite{KOT01}}$.

\subsection{Kalman Filter Approaches}
The Kalman Filter (KF) and its extended version (EKF) are perhaps the most popular prediction approaches employed in a variety of tracking applications$^{\cite{RIS04},\cite{AND05},\cite{KAL60},\cite{JAZ70}}$.  Both, the KF and the EKF, rely on a first order Taylor series approximation of the nonlinear state space model of the system, and assume that process and measurement noises are identically and independently Gaussian distributed.  These assumptions allow the filters to be implemented in a relatively simple recursive way in terms of a {\em state prediction step} and an {\em optimal estimate update} or {\em correction step} as shown in Algorithm~\ref{EKF_ALG}.  However, the main disadvantage of such filtering approaches is their inability to handle systems with non-Gaussian noises and highly nonlinear dynamics, often yielding diverging predictions of the system's state.  This has lead to the recent development of the so-called Sigma-Point Kalman Filters (SPKF)$^{\cite{MER04A}}$, which include the UKF$^{\cite{JUL97}}$, CDKF$^{\cite{ITO00}}$, DD1$^{\cite{NOR00}}$, DD2$^{\cite{NOR00}}$, SRCDKF$^{\cite{MER04A}}$ and SRUKF$^{\cite{MER01}}$ approaches.  Some examples of their applications can be found in references ${\cite{MER04A}}$ and ${\cite{KOT01}}$.  

The SPKF filters do not require the calculation of the system's Jacobians $F_x$, $F_u$, $F_v$ (see Algorithm~\ref{EKF_ALG}), instead samples (sigma points) are generated in a deterministic fashion so that some properties (such as first and second statistical moments) match those of the prior distribution.  Each point undergoes a transformation determined by the system's model equations or some approximation of the model.  The properties of the posterior distribution are obtained from the properties of the transformed sigma points.  The transformation in Julier's UKF is given directly by the model equations, while Ito's CDKF and N{\o}rgaard's DD2 employ a second order Stirling polynomial interpolation formula to approximate the model equations, thereby better retaining second order properties of the samples than the standard EKF and with a potentially smaller computational cost than that of the UKF.  The difference between the DD1 and DD2 filters is in that the former simplifies the model approximation by using a first order Stirling polynomial interpolation formula instead of a second order one.  This simplification allows for some computational time reductions at the expense of accuracy.  Ito's CDKF approach and N{\o}rgaard's DD1, DD2 approaches were developed independently, but are conceptually the same with minor implementation differences.  Later Van der Merwe developed the SRCDKF and SRUKF approaches, suggesting the use of the QR and Cholesky decomposition to compute the square root of the state covariance matrix $P_{k}$.  The square root of $P_{k}$ is required to generate the sigma points in the CDKF, DD1, DD2 or UKF approaches and is one of the most computationally expensive steps of the SPKF filters.  Hence, SRCDKF and SRUKF aim at improving the speed and numerical stability of the CDKF and UKF by placing special attention to the efficient calculation of the square root of the state covariance matrix.   

It must be stressed that no random sampling is used by the SPKF approaches, despite their resemblance to a Monte Carlo method.  Thus only a small number of samples are required to propagate means and covariances.  Because of this, SPKF filters have much smaller computational costs than filters based on Monte Carlo methods$^{\cite{MER04A}}$.  On the other hand, SPKF filters should yield smaller estimation errors than the EKF because of their better approximation of the system's model equations and second order statistical moments. Much like the KF or EKF, the SPKF approaches cannot handle non-Gaussian noises effectively.  The Particle Filters discussed next may overcome this disadvantage.

\restylealgo{ruled}
\setlength{\algomargin}{.8em}
\SetKwInput{KwInit}{Initial Asumptions/Conditions}
\SetKwInput{KwIter}{Iteration k}
\SetKwInput{KwPStp}{1. Prediction Step}
\SetKwInput{KwCStp}{2. Correction Step}
\setalcaphskip{0.3em}
\dontprintsemicolon 
\SetAlgoSkip{}
\begin{algorithm}[H]
\caption{Extended Kalman Filter (EKF)}\label{EKF_ALG}
  \SetLine
  \KwInit{\;
\begin{tabular}{ll}
Initial state estimate: & $\hat{x}_{0}=E\left[x_{0}\right]$\\
Initial state covariance: & $P_{0}=E\left[\left(x_{0}-\hat{x}_{0}\right)\left(x_{0}-\hat{x}_{0}\right)^T\right]$\\
Process noise covariance: & $Q_{0}=E\left[\left(v_{0}-\bar{v}_{0}\right)\left(v_{0}-\bar{v}_{0}\right)^T\right]$\\
Measurement noise covariace: &$R_{0}=E\left[\left(w_{0}-\bar{w}_{0}\right)\left(w_{0}-\hat{w}_{0}\right)^T\right]$
\end{tabular} \;
}
\BlankLine
  \KwIter{For $k=1,2,\ldots,\infty$\;
    \KwPStp{ \;
\Indp
\begin{itemize}
\item Compute the process model Jacobians \; 
${F_x}_k=\nabla_{x}f|_{\left(\hat{x}_{k-1},u_{k-1},\bar{v}_{k-1}\right)}$, \ \ \
${F_u}_k=\nabla_{u}f|_{\left(\hat{x}_{k-1},u_{k-1},\bar{v}_{k-1}\right)}$, \ \ \
${F_v}_k=\nabla_{v}f|_{\left(\hat{x}_{k-1},u_{k-1},\bar{v}_{k-1}\right)}$, \;
\item Compute the predicted state mean and covariance \; 
\begin{tabular}{ll}
State prediction: &$\hat{x}^{-}_{k}=f\left(\hat{x}_{k-1},u_{k-1},\bar{v}_{k-1}\right)$\\
State covariance prediction: & $P_{k}^{-}={F_x}_k P_{k-1} {{F_x}_k}^T+{F_v}_k Q_{k} {F_v}_k^{T}$
\end{tabular} \;
\end{itemize}
\BlankLine
\Indm
    \KwCStp{ \;
\Indp
\begin{itemize}
\item Compute the observation model Jacobians\;
${H_x}_k=\nabla_{x}h|_{\left(\hat{x}_{k}^-,\bar{w}_k\right)}$, \ \ \ 
${H_w}_k=\nabla_{w}h|_{\left(\hat{x}_{k}^-,\bar{w}_k\right)}$\;
\item Update the state estimate with the lastest observation (measurement update)\; 
\begin{tabular}{ll}
Kalman gain: &$K_{k}=P_{k}^{-}{H_x}_k^{T}\left({H_x}_k P_k^{-} {H_x}_k^{T}+{H_w}_k R_{k} {H_w}_k^{T}\right)^{-1}$\\
State estimate: & $\hat{x}_{k}=\hat{x}^{-}_{k}+K_{k}\left[y_{k}-h\left(\hat{x}^{-}_{k},\bar{w}_k\right)\right]$ \\
State covariance: &$P_{k}=\left(I-K_{k}{H_x}_k\right)P_{k}^{-}$
\end{tabular} \;
\end{itemize}
       }
    }
}
\end{algorithm}

\subsection{Particle Filter Approaches}
Particle Filters (PF) belong to the class of Sequential Monte Carlo (SMC) methods, which rely on the idea of recursively computing the posterior probability distributions using concepts of importance sampling and the approximation of distributions by discrete random measures (particles) from the space of unknown states$^{\cite{RIS04},\cite{DOU00},\cite{MER00},\cite{MER03}}$.  The application of SMC methods dates back to the 1950's$^{\cite{MET49}}$, even before the development of the KF, however it has not been until the last decade that PF have received significant attention$^{\cite{RIS04},\cite{DOU00},\cite{GOR93}}$, mainly because the computational complexity involved demands adequate computing resources that did not exist before. 

The main steps of the PF computation are shown in Algorithm~\ref{PF_ALG}.  Whereas KF/EKF and SPKF methods assume that noises are Gaussian and independently identically distributed (i.i.d.) in order to simplify the formulation of the optimal recursive Bayesian estimator, PF methods make no assumptions on the form of the probability densities.  Moreover, the particles (samples) interact and thus PF methods may take into account the statistical dependence of the samples.  Therefore, PF methods present a greater potential to successfully handle strongly nonlinear, non-Gaussian processes.  The shortcoming of SMC approaches is the high computational effort that they require  due to the large number of samples needed to obtain accurate state estimates.  The number of samples may grow exponentially with respect to the dimension of the state space, especially for the standard PF$^{\cite{RIS04}}$.  To overcome this disadvantage, several ``hybrid'' methods have been proposed to generate the particles and weights based on KF techniques and its extensions.  These hybrid methods allow to obtain accurate state estimates without having to generate a large number of samples, but yet again do not perform well if noises do not conform to the Gaussian assumption.

A method to deal with non-Gaussian noises by discarding only those measurements that are deemed unlikely to occur under the Gaussian noise assumption has been proposed by Browning et al.$^{\cite{BRO02}}$. The method is referred to as Improbability Filter (IF), and can be employed in combination with the KF or the PF methods that cannot handle non-Gaussian noises properly.  The next section briefly explains the IF because of its usefulness towards improving the estimates' accuracy.

\restylealgo{ruled}
\setlength{\algomargin}{.8em}
\SetKwInput{KwInit}{Initialization}
\SetKwInput{KwIter}{Iteration k}
\SetKwInput{KwIS}{1. Importance Sampling Step}
\SetKwInput{KwSS}{2. Selection Step (Resampling)}
\SetKwInput{KwO}{3. Output}
\setalcaphskip{0.3em}
\dontprintsemicolon 
\SetAlgoSkip{}
\begin{algorithm}[H]
\caption{Particle Filter (PF)}\label{PF_ALG}
\SetLine
\KwInit{\;
\Indp
For $i=1,2,\ldots,N$ draw samples (particles) $x^{(i)}_0$ from the prior $p(x_{0})$. \;
}
\BlankLine
  \KwIter{For $k=1,2,\ldots,\infty$ \;
    \KwIS{ \;
\Indp
For $i=1,2,\ldots,N$:
\begin{itemize}
\item Draw samples $x^{(i)}_k \sim p\left(x_{k}|x_{k-1}^{(i)}\right)
\leftarrow f\left (x_{k-1}^{(i)},u_{k-1},v_{k-1}^{(i)}\right )$.\;
\item Evaluate the importance weights up to a normalizing constant:  
$\tilde{w}_{k}^{(i)}=w_{k-1}^{(i)}p\left(y_{k}|x_{k}^{(i)}\right)=
w_{k-1}^{(i)}p_w\left(y_{k}-h\left (x_{k}^{(i)}\right)\right)$, asuming $y_k=h(x_k)+w_k$, $w_k\sim p_w$. \;
\item Normalize the importance weights:  $w_{k}^{(i)}=\frac{\tilde{w}_{k}^{(i)}}{\sum^{N}_{j=1}{\tilde{w}_{k}^{(i)}}}$. \; 
\end{itemize}
\BlankLine
\Indm
    \KwSS{ \;
\Indp
\begin{itemize}
\item Multiply/suppress samples $x_{k}^{(i)}$ with high/low importance weights $w_{k}^{(i)}$ to obtain $N$ random samples approximately distributed according to $p(x_{k}|Y_{k})$. \; 
\item For $i=1,2,\ldots,N$ set $w_{k}^{(i)}=N^{-1}$.\; 
\end{itemize}

\BlankLine
\Indm
    \KwO{ \;
\Indp
\begin{itemize}
\item Using the samples, the state estimate is calculated as the samples expected value\; 
 $\hat{x}_{k}=E\left( x_{k}|Y_{k} \right) \approx \frac{1}{N}\sum^{N}_{i=1}{x_{k}^{(i)}}$. \; 
\end{itemize}
       }
       }
    }
}
\end{algorithm}

\subsection{Improbability Filter}
Browning's Improbability Filter (IF) was proposed as a simple technique to prevent abnormal observations from degrading the state estimates while preserving a good filter response.  The implicit assumption of this filter is that the measurements are locally Gaussian with mean given by the predicted measurement, $\hat{y}_k=h_{k}(\hat{x}_{k},\bar{w}_k)$, and measurement covariance matrix $S_{k}={H_x}_k P_k {H_x}_k^{T}+{H_w}_k R_{k} {H_w}_k^{T}$.  With this assumption, a criterion to discard {\em unlikely} measurements is to reject those observations $y_k$ whose conditional probability of occurrence given $\hat{y}_k$ and $S_k$: 
\begin{eqnarray}
P\left( y_k|\hat{y}_{k},\,S_{k} \right) =\frac{1}{\left( 2\pi
\left\vert S_{k}\right\vert \right)^{\frac{n}{2}}}\cdot e^{-\frac{1}{2}
\left(y_k-\hat{y}_{k}\right) ^{T}S_{k}^{-1}\left(
y_k-\hat{y}_{k}\right) }\la{IF}
\end{eqnarray}
falls below certain threshold.  An equivalent criterion is to use the Mahalanobis distance
\begin{eqnarray}
d_{M}(y_k;\hat{y}_k,S_k)=\left(y_k-\hat{y}_k\right)
^{T}S_{k}^{-1}\left(y_k-\hat{y}_k\right)\la{MD}
\end{eqnarray}
in order to discard those observations $y_k$ whose distance with respect to the predicted measurement $\hat{y}_k$ exceeds a given threshold. If an observation is discarded, the Improbability Filter assumes that the predicted measurement is the correct measurement and no state update is performed. However, state covariances are updated through the dynamic update equation. If measurements are repeatedly rejected, the state covariances $P_{k}$ will grow until observations become likely and are accepted. Furthermore, the large covariances will make the state estimate to converge immediately according to the new observations.

\subsection{Differences between KF, PF and IF}
Concisely, the difference between KF and PF approaches lies in the fact that KFs assume the noises have a Gaussian distribution and that the model can be reasonably approximated by a linear Taylor series approximation, while PF make no specific assumption about the distribution of the process or measurement noises, nor they necessarily make any approximation of the model equations.  This difference is reflected by the fact that the EKF uses the Jacobians of the system equations~\re{state_space_model} and~\re{measurement_model}, (denoted by $\nabla$ in Algorithm~\ref{EKF_ALG}), to propagate the covariances that define the probability of the optimal state estimate $\hat{x}_k$, while the PF approaches propagate samples $x_k^{(i)}$ with transition probability $p\left (x_k|x_{k-1}^{(i)}\right )$ defined by the model equation~\re{state_space_model} and the known statistics of the process noise $v_{k-1}$.  In simple terms, the PF requires generating a number $N$ of possible states which, after undergoing a transformation defined by the model equation~\re{state_space_model}, allow to reconstruct the posterior probability~\re{PROB1} using importance weights that are calculated in terms of the measurements likelihood $p(y_k|x_k^{(i)})$; (see Algorithm~\ref{PF_ALG}).  The KF filters propagate only one state, whose probability of being the actual state is given by a Gaussian distribution with mean $\hat{x}_k$ and covariance $P_k$ resulting from the update equations; (see Algorithm~\ref{EKF_ALG}).  It is to be noted that the UKF algorithm involves analogous steps to those of Algorithm~\ref{EKF_ALG}, but does not require the computation of the Jacobians because the model equation~\re{state_space_model} is employed to deterministically generate some samples that preserve first and second order statistics.  Other SPKF algorithms, such as CDKF, DD1, DD2, SRCDKF, SRUKF are similar to UKF in the sense that the statistics information is also carried by a number of particles that are deterministically chosen, but also rely on first or second order model approximations. 

From an implementation perspective, the computational complexity of the KF approaches per recursion is of order $\mathcal{O}(n^3+m^3+2n^2m+2nm^2+c_0)$, where $n$ and $m$ are the dimensions of the state space model and the measurement model, respectively$^{\cite{GOR97}}$, and $c_0$ is the cost of computing the gradients or model equations.  On the other hand, the PF methods require have a computational complexity which is roughly of order $\mathcal{O}(\frac{4}{3}n^3+2n^2)+N\cdot\mathcal{O}(4n^2+n+c_1+c_2+nc_3)$, where $n$ is the dimension of the state space, $N$ is the number of particles, $c_1$ is the cost of the calculation of the likelihood function, $c_2$ is the cost of the resampling step and $c_3$ is the complexity of the random number generator$^{\cite{KAR05}}$.   Assuming $n=m$, for large $n$ the complexity of KFs is roughly $\mathcal{O}(6n^3)$, while that of PFs is $\mathcal{O}(4Nn^2)+\mathcal{O}(\frac{4}{3}n^3)>>\mathcal{O}(6n^3)$ since $N>>n$.  It is hard to make an exact comparison because the complexity  may vary depending on the level of nonlinearity exhibited by the system equations and whether first or second order approximations are employed. However, it is to be noted that the main costs of the KF arise from the computation of the Riccati equation for matrix $P^-_{k}$ due to the matrix multiplication $F_{x_k}P_{k-1}$, which is $\mathcal{O}(2n^3)$.  The general PF does not involve the matrix times matrix multiplication, instead only $\mathcal{O}(n^2)$ matrix times vector multiplications $F_{x_k}x_k$ are required to generate the samples $N$ times, with $N$ several times larger than $n$.  It is also worth pointing that even if the relationship between the theoretical complexity (expressed in terms of the number of floating-point operations) and the actual computation time is not straight forward because of the influence of memory locality and cache boundaries, the results nonetheless confirm that PF-based approachs are computationally more time consuming that KF-based methods.

Finally, it must be noticed that the IF is not a recursive Bayesian estimator, but it exploits the assumption that the measurement noises are Gaussian to discard those measurements which deviate from the expected measurement and whose probability in the Gaussian sense is low given the knowledge of the updated measurement covariance.  Using the IF in conjunction with KF, EKF or other SPKF approaches makes sense because the IF should be able to reject measurements that are not Gaussian and that would otherwise deteriorate the prediction and update of the covariances.  Since the PF methods do not require to assume Gaussian noises nor linear models, using the IF with the PF approaches should not offer any benefit and just increase the computational cost.

\section{Omni-directional Robot Model}\label{sec:model}
The performance of the filtering techniques mentioned in the previous section, as applied to the position tracking and prediction problem, is evaluated using a three-wheel omni-directional robot as test platform.  It is worth pointing out that the particular choice of model does not limit the validity of the performance tests.  If a robot with a different configuration is employed, then the model equations should be updated accordingly.  The choice of the three-wheel configuration is based on the fact that: (i) it possesses a nonlinear dynamics which offers an adequate challenge to the filtering techniques, (ii) it is simple to control as there are less actuators, and (iii) it is the type of robot that is currently being employed by the PUC's F-180 RoboCup team.

The kinematic and dynamic model is presented next following the work by Liu et al.$^{\cite{LIU03}}$; for a detailed and rigorous derivation the reader is referred to the book by Angeles$^{\cite{ANG06}}$.  For clarity of exposition, the notation employed in the formulation of the robot's model is summarized in table~\ref{tab:symbol}.  The model equations are derived by expressing velocities and forces acting on the robot in terms of their components along the local coordinate system, $\left \{\mathbf{x^r},\mathbf{y^r}\right \}$, which is fixed to the robot's body as illustrated in fig.~\ref{fig:robot_lcs}.  Simple vector rotation operations allow to rewrite the robot's kinematic and dynamic equations in terms of the global coordinate system, $\left \{\mathbf{x^w},\mathbf{y^w}\right \}$, which is fixed to the court as shown in fig.~\ref{fig:robot_gcs}.

\begin{table}[htbp]
\begin{center}
\small
\begin{tabular}{cccc}
\hline
\multicolumn{2}{c}{\sc Symbol}
& \multicolumn{2}{c}{\sc Description}\\
\hline
\multicolumn{4}{l}{\bf Robot Parameters}\\
$\mathbf{o}_1$, $\mathbf{o}_2$, $\mathbf{o}_3$ & & \multicolumn{2}{c}{rear, front-right and front-left wheels}\\
$m$       & & \multicolumn{2}{c}{robot's mass}\\
$I_0$     & & \multicolumn{2}{c}{robot's inertia moment}\\
$r$       & & \multicolumn{2}{c}{wheel radius}\\
$L$       & & \multicolumn{2}{c}{body radius}\\
$N$       & & \multicolumn{2}{c}{gear ratio}\\
$J_0$     & & \multicolumn{2}{c}{motor armature and wheel inertia moment}\\ 
$K_T$, $K_G$, $R_a$, $L_a$ &  &
              \multicolumn{2}{c}{motor constants (torque, EMF, 
                                 resistance and inductance)}\\

\multicolumn{4}{l}{\bf Robot Local Variables (Body Frame)}\\
$\mathbf{x^r}, \mathbf{y^r}$  & & \multicolumn{2}{c}{axes of the local coordinate frame fixed to the robot's body}\\
$F_{\mathbf{x^r}}, F_{\mathbf{y^r}}$ & & \multicolumn{2}{c}{traction forces acting on the robot along $\mathbf{x^r}$ and $\mathbf{y^r}$}\\
$\dot{x}^r, \dot{y}^r$ & & \multicolumn{2}{c}{robot velocities along $\mathbf{x^r}$ and $\mathbf{y^r}$}\\
$\tau_\theta$         & & \multicolumn{2}{c}{body rotation torque}\\
$\dot{\theta}$ & & \multicolumn{2}{c}{robot's heading rate of change}\\
$
f_{1}, f_{2}, f_{3}
$
        & & \multicolumn{2}{c}{traction forces of the wheels}\\
$
\dot{\phi}_1, \dot{\phi}_{2}, \dot{\phi}_3
$
          & & \multicolumn{2}{c}{angular velocities of the motors}\\
$
u_{1}, u_{2}, u_{3}
$
        & & \multicolumn{2}{c}{input voltages of the motors}\\
\multicolumn{4}{l}{\bf Robot World Variables (Court Frame)}\\
$\mathbf{x^w}, \mathbf{y^w}$  & & \multicolumn{2}{c}{axes of the global coordinate frame fixed to the court}\\
$\mathbf{p}=(x^w, y^w)$ & & \multicolumn{2}{c}{robot location}\\
$\theta$   & & \multicolumn{2}{c}{\parbox[t]{.72\textwidth}{\centering robot heading\\ (defined as the angle between the robot's $\mathbf{y^r}$ axis and the global $\mathbf{x^w}$ axis)}}\\
\hline
\end{tabular}
\end{center}
\caption{Notation for the model equations of the three-wheel robot.}
\label{tab:symbol}
\end{table}

 \begin{figure}[htbp]
\begin{center}
\includegraphics[scale=1]{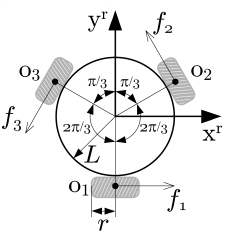}\\
\end{center}
\vspace*{-1.5em}
\caption{Robot coordinate frame $\left \{\mathbf{x^r},\mathbf{x^r}\right \}$ fixed to the robot's body (local coordiantes) and the traction forces $f_1$, $f_2$ and $f_3$ of the wheels.}
\label{fig:robot_lcs}
\end{figure}

\begin{figure}[htbp]
\begin{center}
\includegraphics[scale=1]{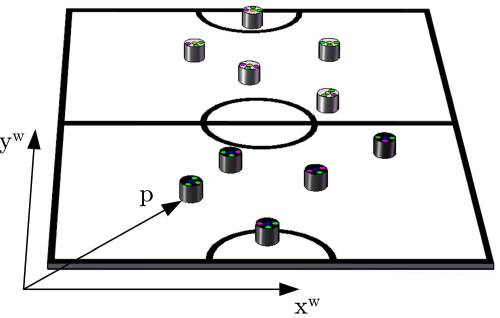}
\end{center}
\vspace*{-1.5em}
\caption{World coordinate frame $\left \{\mathbf{x^w},\mathbf{y^w}\right \}$ fixed to the court (global coordinates) and two teams of RoboCup F-180 robots.}
\label{fig:robot_gcs}
\end{figure}

\begin{figure}[htb]
\begin{center}
\includegraphics[scale=1]{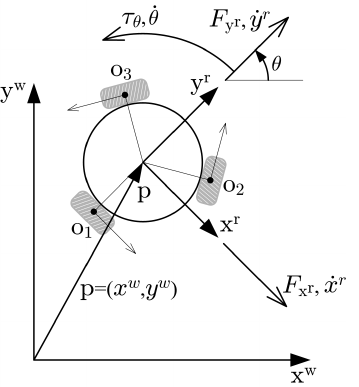}
\end{center}
\vspace*{-1.5em}
\caption{Forces acting on the robot and velocity components (see Table~\ref{tab:symbol} for notation details).}
\label{fig:robot_forces}
\end{figure}

The following simplifying assumptions regarding the system have been made:
\begin{enumerate}
\item The robot body only undergoes planar motion, and there is no wheel slippage, nor skidding.
\item Dynamic friction on the robot is negligible and only viscous friction in the motors and bearings is considered.
\item The mass and moment of inertia of the wheels about the $\mathbf{z^r}$ axis perpendicular to $\mathbf{x^r}$ and $\mathbf{y^r}$ have been lumped into a mass term $m$ and a general inertia $I_0$.
\end{enumerate}
By considering these standard assumptions, additional uncertainty about the true model parameters is introduced.  However, this uncertainty is overcome by the estimation strategy.  The increased uncertainty poses an additional {\em desirable challenge} to the algorithms that are to be evaluated in the following section.

\subsection{Robot Kinematics}
Defining the matrices
\begin{eqnarray}
P_0\ \stackrel{def}{=}\  
\left [\begin{array}{ccc}
 \sin(\theta) & \cos(\theta) & 0\\
-\cos(\theta) & \sin(\theta) & 0\\
0 & 0 & 1
\end{array}\right ],\ \ \ 
P_1\ \stackrel{def}{=}\  
\left [\begin{array}{ccc}
1 & 0 & L\\
-\sin(\psi) &  \cos(\psi) & L\\
-\sin(\psi) & -\cos(\psi) & L
\end{array}\right ],\ \text{with } \psi=\frac{\pi}{6}\nonumber
\end{eqnarray}
it is easy to verify from the geometry of the robot shown in fig.~\ref{fig:robot_lcs} and fig.~\ref{fig:robot_forces}, that the robot's velocity components $\left [\dot{x}^r,\dot{y}^r,\dot{\theta}\right ]$ in the local frame $\left \{\mathbf{x^r},\mathbf{y^r}\right \}$ are related to the wheel velocities according to:
\begin{eqnarray}
P_1\mathbf{v^r}&=&\frac{r}{N}\dot{\boldsymbol{\phi}}\la{kin_lcs}
\end{eqnarray}
where $\mathbf{v^r}=\left [\dot{x}^r,\dot{y}^r,\dot{\theta}\right ]^T$, $\dot{\boldsymbol{\phi}}=\left [\dot{\phi}_1,\dot{\phi}_2,\dot{\phi}_3\right ]^T$.  The kinematics equation defining robot's velocity components vector in the global coordinates, $\mathbf{v^w}$, as a function of the wheels' angular velocities is then given by:
\begin{eqnarray}
\mathbf{v^w}=P_0\mathbf{v^r}=\frac{r}{N}P_0P_1^{-1}\dot{\boldsymbol{\phi}}\la{kin_gcs}
\end{eqnarray}

\subsection{Robot Dynamics}
Defining the robot's generalized inertia matrix as:
\begin{eqnarray}
I & \stackrel{def}{=} &
\left[\begin{array}{ccc}
m & 0 & 0\\
0 & m & 0\\
0 & 0 & I_0
\end{array}\right ]\nonumber
\end{eqnarray}
and applying Newton's Law in the body frame
\begin{eqnarray}
I\dot{\mathbf{v}}^\mathbf{r} &=&
I\dot{\theta}\mathbf{z^r}\times \mathbf{v_p^r}+P_1^T\mathbf{f}\la{dyn_lcs}
\end{eqnarray}
where $\mathbf{v_p^r}=\dot{x}^r\mathbf{x^r}+\dot{y}^r\mathbf{y^r}$ is the velocity of the robot center of mass $\mathbf{p}$ in the $\mathbf{x^r}$ and $\mathbf{y^r}$ directions, and $\mathbf{f}=[f_1,f_2,f_3]^T$ is the vector of net traction forces of the wheels.

The dynamics of the DC motors is described by the equations:
\begin{eqnarray}
J_0\ddot{\boldsymbol{\phi}}&=&K_T\mathbf{i}_a-b_0\dot{\boldsymbol{\phi}}-\frac{r}{N}\mathbf{f}\la{mot_dyn}\\
\mathbf{u}&=&R_a\mathbf{i}_a+L_a\frac{d\mathbf{i}_a}{dt}+K_G\dot{\boldsymbol{\phi}}\la{mot_elec}
\end{eqnarray}
where $\mathbf{i}_a=\left [{i_a}_1,{i_a}_2,{i_a}_3\right ]$ is the vector of armature currents, $\mathbf{u}=\left [u_1,u_2,u_3\right ]$ is the vector of armature voltages.  Considering that the time constant of the electrical equation is very small compared to that of the mechanical equation, it is possible to neglect the dynamics of the motors' electrical circuits, i.e. $d\mathbf{i}_a/dt=0$, $\mathbf{i}_a=(\mathbf{u}-K_G\dot{\boldsymbol{\phi}})/R_a$, and therefore it is possible to rewrite the motors' mechanical equation~\re{mot_dyn} as:
\begin{eqnarray}
J_0\ddot{\boldsymbol{\phi}}&=&K_T\frac{\mathbf{u}-K_G\dot{\boldsymbol{\phi}}}{R_a}
-b_0\dot{\boldsymbol{\phi}}-\frac{r}{N}\mathbf{f}\la{mot_dyn1}
\end{eqnarray}
From~\re{kin_lcs}, $\dot{\boldsymbol{\phi}}=\frac{N}{r}P_1\mathbf{v^r}$ and $\ddot{\boldsymbol{\phi}}=\frac{N}{r}P_1\dot{\mathbf{v}}^\mathbf{r}$, which replaced in~\re{mot_dyn1} yield:
\begin{eqnarray}
\mathbf{f}&=&
-a_1 P_1\dot{\mathbf{v}}^\mathbf{r}
+a_2 P_1\mathbf{v^r}
+a_3 \mathbf{u}
\la{mot_dyn2}
\end{eqnarray} 
where,
\begin{eqnarray}
a_1=J_0\frac{N^2}{r^2},\ a_2=-\frac{N^2}{r^2} \left (\frac{K_TK_G}{R_a}+b_0 \right ),\ a_3=\frac{K_TN}{R_a r}\nonumber
\end{eqnarray}

Finally, replacing~\re{mot_dyn2} in~\re{dyn_lcs} results in the following equation for the robot's dynamics:
\begin{eqnarray}
H \dot{\mathbf{v}}^\mathbf{r} & = &
\dot{\theta}\mathbf{z^r}\times \mathbf{v_p^r} 
+a_2 I^{-1}P_1^T P_1\mathbf{v^r}
+a_3 I^{-1}P_1^T\mathbf{u}
\la{dyn_lcs1}
\end{eqnarray}
with $H\stackrel{def}{=}\left (\mathbb{I}_3+a_1 I^{-1}P_1^T P_1\right)$.  Here, $\mathbb{I}_{3\times 3}$ denotes the $3\times 3$ identity matrix.  In vector form, equation~\re{dyn_lcs1} can be written as:
\begin{eqnarray}
H
\left [\begin{array}{c}
\ddot{x}^r\\\ddot{y}^r\\\ddot{\theta}
\end{array}\right ]
& = &
\left [\begin{array}{c}
\dot{\theta}\dot{y}^r\\-\dot{\theta}\dot{x}^r\\0
\end{array}\right ]
+a_2 
I^{-1}P_1^T P_1
\left [\begin{array}{c}
\dot{x}^r\\\dot{y}^r\\\dot{\theta}
\end{array}\right ]
+a_3 
I^{-1}P_1^T
\left [\begin{array}{c}
u_1\\u_2\\u_3
\end{array}\right ]
\la{dyn_lcs2}
\end{eqnarray}

\subsection{Linear Discrete-Time State Space Model}\label{sec:dtlssm}
In order to formulate a discrete-time model for predicting the robot's motion, let $x_k\stackrel{def}{=}\left  [x_k^w,y_k^w,\theta_k,\dot{x}_k^r,\dot{y}_k^r,\dot{\theta}_k\right ]^T$ be the state vector that describes the robot's motion at discrete time instants $t_k=kT$, $k=0,1,2,\ldots$, for some sampling period $T>0$.  It is assumed that the state of the robot is affected by an i.i.d. zero-mean disturbance $v_k$ with (constant) covariance $Q_k=E\left (v_k v_k^T\right )$.
Linearization of the kinematic and dynamic equations~\re{kin_gcs} and~\re{dyn_lcs2} about a nominal state $x_k^e\stackrel{def}{=}\left  [x_k^w,y_k^w,\theta_k,0,0,0\right ]^T$, followed by discretization with sampling period $T$ allows to obtain the following difference equation defining the evolution of the robot's state:
\begin{eqnarray}
{x}_k=A_{k-1}{x}_{k-1}+B u_{k-1}+v_{k-1}
\la{DE1}
\end{eqnarray}
with 
\begin{eqnarray}
A_j=\left [ 
\begin{array}{cc}
\mathbb{I}_{3\times 3} & P_0\left( \theta _{j}\right) T \\ 
\mathbf{0}_{3\times 3} & \mathbb{I}_{3\times 3}+A_{22} T 
\end{array}
\right ],\  
A_{22}=
\left [
\begin{array}{ccc}
\frac{3 a_2}{2m+3a_1} & 0 & 0\\
0 & \frac{3 a_2}{2m+3a_1} & 0\\
0 & 0 & \frac{3 a_2}{I_0/L^2+3a_1}
\end{array}
\right ]
\la{DEM1}
\end{eqnarray}
and
\begin{eqnarray}
B=\left [
\begin{array}{c}
\mathbf{0}_{3\times 3}\\
B_2 T
\end{array}
\right ],\  
B_2=\frac{a_3}{2m}
\left [
\begin{array}{ccc}
2 & -1 & -1\\
0 & -\sqrt{3} & -\sqrt{3}\\
\frac{2mL}{I_0} & \frac{2mL}{I_0} & \frac{2mL}{I_0} 
\end{array}
\right ].
\end{eqnarray}

Here $\mathbf{0}_{3\times 3}$ represents the $3\times 3$ matrix of zeroes, while $P_0(\theta_j)$ is employed to emphasize the dependence of the rotation matrix $P_0$ on the value of the heading angle $\theta_j$.  It is also important to point out that the inputs vector is actually defined as the vector of delayed armature voltages as $u_k=[V_{1_{k-l}}, V_{2_{k-l}}, V_{3_{k-l}}]^T$, where $V_{i_{k-l}}$, $i=1,2,3$, is the armature voltage of the $i$-th motor delayed $l$ sampling periods.  The delay $l$ accounts for the system's latency, mostly due to the delay arising in the image acquisition and segmentation process that is employed to obtain each measurement of the robot's position and heading.

The measurement model considers an observation vector given by:
\begin{eqnarray}
{y}_{k}=h\left({x}_{k},w_k\right) &=&
\left [ 
{x}^w_k+w_{1_k}, {y}^w_k+w_{2_k}, {\theta}_k+w_{3_k}
\right ]^T
\la{MESE}
\end{eqnarray}
where $w_{i_k}$, $i=1,2,3$, are the measurement noises, which are assumed to be the mixture of a Gaussian and a non-Gaussian random process, i.e. $w_{i_k}=w^g_{i_k}+w^n_{i_k}$, $i=1,2,3$, where $w^g_{i_k}$ and $w^n_{i_k}$ are the Gaussian and non-Gaussian components, respectively.  The Gaussian process $w^g_k=\left [w^g_{1_k},w^g_{2_k},w^g_{3_k}\right ]^T$ is assumed to be i.i.d. for all $k\in\mathbb{N}$, with zero-mean and (constant) covariance matrix $R_k=E(w^g_k{w^g_k}^T)$.  The non-Gaussian process $w^n_k=\left [w^n_{1_k},w^n_{2_k},w^n_{3_k}\right ]^T$ takes values from a uniform distribution with a probability of occurrence at each instant $k$ defined by a Bernoulli distribution.

The discrete-time state-space model defined by equations~\re{DE1} and~\re{MESE} allows to generate the ideal and noisy trajectories for typical input commands applied to the robots during their operation.  These trajectories will serve in the evaluation of the different filtering and prediction approaches as explained in the next sections.

\section{Testing Platform and Methodology}\label{sec:methodology}
This section briefly describes the general architecture of the RoboCup F-180 team of robots employed as testbed, the filter implementation and the testing methodology.

\subsection{RoboCup F-180 System Architecture}
The test bench system employed to assess the prediction capability of the filters is based on the simulation of motion tracking under noisy conditions of the three-wheel omnidirectional robot presented in the previous section.  The best performing method is verified experimentally using real robots of the PUC RoboCup team.  The team of five robots is controlled by a system whose architecture is illustrated in Fig.~\ref{fig:predsys}.  As shown in  Fig.~\ref{fig:predsys}, the vision system implemented in a standard PC receives images from a Samsung SCD107 digital video camera with a Sony VCL-0625 wide angle lens via an IEEE 1394 Firewire connection.  The vision system extracts the position $({x}^w,{y}^w)$ and heading ${\theta}$ of the robots using a standard color segmentation process.  These measurements are then passed to the planning and control PC, which is responsible for filtering the measurements, predicting future states and generating motion commands aimed at driving certain robots to the ball and scoring.  The filter is required to reduce robot misdetection and false positives that may occur due to fast motion and color confusion.  The prediction of states is fundamental due to the unavoidable delays arising mainly from actuator response times, the computation time required by the vision and control system, as well as the communication bandwidth that limits data throughput.  All these delays can add up to 160 ms, which are equivalent to missing 5 video frames for a standard 30 fps video camera.  The total delay is non-negligible, especially when a robot reaches its maximum velocity of 1.25 m/s, as this implies that the robot's displacement between one frame and the next processed one 160 ms later amounts to approximately 20 cm.  This position difference is quite significant considering that the robot maximum diameter is 18 cm.  For this reason, the planning and control system must receive an estimate of the future position of the robot corresponding to four frames time ahead, so that the planning and control system can issue adequate command decisions.

\begin{figure}[htb]
\begin{center}
\includegraphics[scale=0.7]{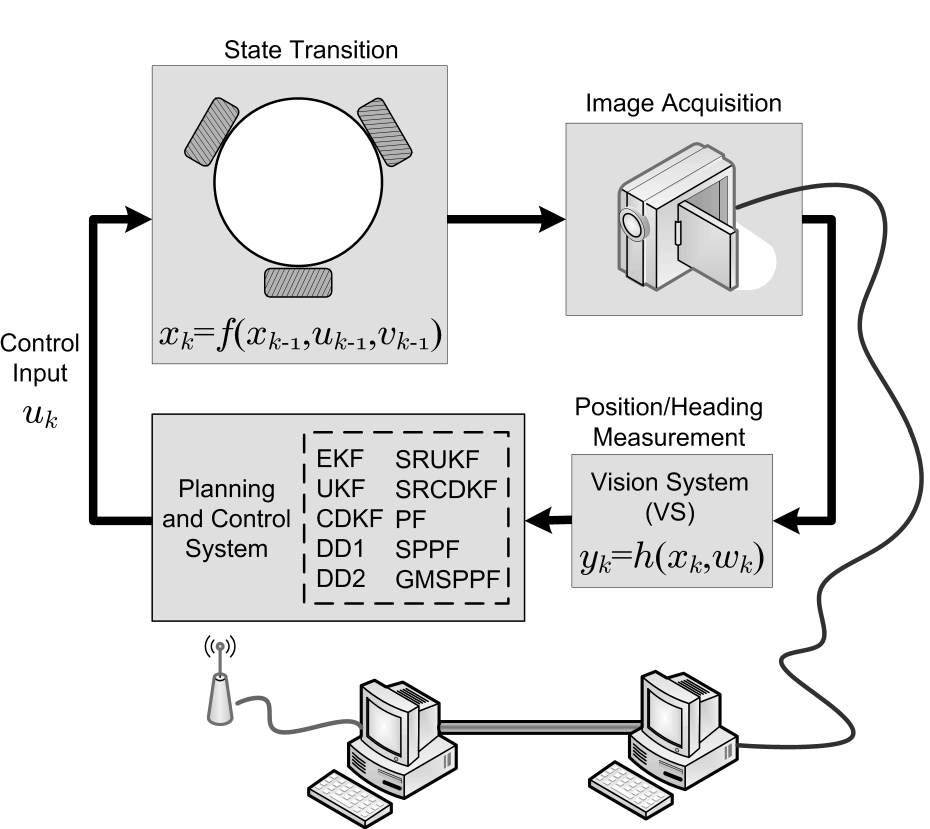}
\end{center}
\vspace*{-1.5em}
\caption{PUC's RoboCup F-180 system architecture.}
\label{fig:predsys}
\end{figure}

\subsection{Filter Implementation}
The filters were implemented using Matlab/Simulink$^\circledR$ and some modified functions of the ReBEL$^{\cite{MER02_REBEL}}$ and Kalmtool$^{\cite{NOR98_KALMTOOL}}$ Matlab$^\circledR$ toolkits designed to facilitate recursive Bayesian estimation and prediction calculations.

The process noise covariance matrix was defined as $Q_k\stackrel{def}{=}\mathrm{diag}\left (\sigma_{\dot{x}^w}^2, \sigma_{\dot{y}^w}^2, \sigma_{\dot{\theta}}^2, \sigma_{\ddot{x}^r}^2, \sigma_{\ddot{y}^r}^2, \sigma_{\ddot{\theta}}^2\right )$, where $\mathrm{diag}(\cdot)$ stands for the diagonal matrix of its arguments.  The standard deviations for the accelerations $\ddot{x}^r, \ddot{y}^r, \ddot{\theta}$ were chosen so that their values reflect the acceleration limits imposed by the motors and wheel slippage$^{\cite{BRO02}}$.  In a similar way, the values of the standard deviations for the velocities $\dot{x}^w, \dot{y}^w, \dot{\theta}$ were defined as the largest change in velocity achievable over a sampling period $T$.  It was verified experimentally, that the maximum robot linear acceleration was 1.5 $m/s^2$, therefore $\sigma_{\ddot{x}^r}=\sigma_{\ddot{y}^r}=2\,m/s^2$ is a reasonable choice to characterize the largest possible acceleration disturbance.  The angular acceleration value is also chosen slightly above the maximum angular acceleration measured, setting $\sigma_{\ddot{\theta}}=2\,rad/s$.  With a sampling period $T=0.0333\ s$, experimental verification lead to the following values for the velocities standard deviations: $\sigma_{\dot{x}^w}=\sigma_{\dot{y}^w}=0.067\,m/s$, $\sigma_{\dot{\theta}}=0.2\,rad/s$. 

Similarly, the Gaussian measurement noise covariance matrix was defined as a diagonal matrix $R_k\stackrel{def}{=}\mathrm{diag}\left (\sigma_{{x}^w}^2, \sigma_{{y}^w}^2, \sigma_{{\theta}}^2\right )$ with $\sigma_{{x}^w}=\sigma_{{y}^w}=0.005\,m$ and $\sigma_{{\theta}}=0.020\,rad$ experimentally determined.  The position and heading measurements also affected by non-Gaussian noise generated from a uniform distribution that may produce an observation corresponding to an apparent {\em instantaneous relocation} of the robot to any position within the court and with a possible sudden change in heading.

The initial value of the state vector is taken to be ${x}_0\stackrel{def}{=}\left  [{x}_0^w,{y}_0^w,{\theta}_0,\dot{{x}}_0^r,\dot{{y}}_0^r,\dot{{\theta}}_0\right ]^T=\left  [\bar{x}_0^w,\bar{y}_0^w,\bar{\theta}_0,0,0,0\right]^T$, where $\bar{x}_0^w,\bar{y}_0^w,\bar{\theta}_0$ denote the measured initial position and heading.  The state covariance matrix is initialized with the value $P_0\stackrel{def}{=}Q_k\cdot T^2$, which corresponds to the largest expected initial position and velocities deviation after one sampling period $T=0.0333\,s$.

\subsection{Testing Methodology}
The ability of the different filters to accurately predict the robot's state at different time horizons is measured in terms of the error between the ideal (noise-free) trajectory and the estimated one.  To this end, simulations were performed using six different reference trajectories, which are shown in Fig.~\ref{fig:diff_traj} and that were generated using the noise-free version of the state space model of section 3.3.  The trajectories comprehend different manoeuvres that are common in RoboCup F-180 competitions, including rectilinear and circular paths with different motion regimes, such as constant acceleration, constant velocity, constant rate of turning or their combination.  The test trajectories are purposely constructed to provide challenging situations in which it is more difficult to track turns or sudden changes in direction.  The trajectories employed here are not intended to be optimal trajectories, although our trajectory planner does solve an optimization problem in real-time to dynamically find the best path that avoids the robots of the other team.

Twenty simulation runs were repeated for each of the six reference trajectories using a new random number generator seed each time.   Figure~\ref{fig:Ntrj} shows four of the twenty simulation runs corresponding to the reference trajectory in the first row, second column of fig.~\ref{fig:diff_traj}. In the noisy trajectories of fig.~\ref{fig:Ntrj}, it is possible to appreciate the non-Gaussian noise in the observations due to erroneous detections, which appear as sudden jumps in the {\em measured} location of the robot.  Each trajectory simulation represents 20 seconds of motion with measurement sampling frequency of 30 Hz (corresponding to the frame-grabber 30 fps acquisition rate), and thus each simulation results in 601 trajectory points.

The root mean squared error (RMSE) of the position $e_p=\sqrt{\frac{1}{N}\sum_{k=1}^{N} (\check{x}^w_k-\hat{x}^w_k)^2+(\check{y}^w_k-\hat{y}^w_k)^2}$ (i.e. Euclidean distance between a reference position $(\check{x}^w,\check{y}^w)$ and the estimated position $(\hat{x}^w,\hat{y}^w)$) and the RMSE of the heading angle $e_\theta=\sqrt{\frac{1}{N}\sum_{k=1}^{N}(\check{\theta}_k-\hat{\theta}_k)^2}$ (where $\check{\theta}_k$ is the reference heading) are calculated for each of the simulation runs, and then averaged to assess the quality of the estimation resulting from the different filters.  With the exception of the Particle Filter, which makes no assumption about Gaussianity of the measurements, all filters were evaluated with and without a cascaded Improbability Filter in order to verify possible improvements achievable with this scheme for rejecting non-Gaussian disturbances.  The performance of each filter is also measured in terms of its computation time, which is expressed in relative terms with respect to the fastest filter.  Expressing computation time in relative terms is more useful than presenting absolute time amounts because absolute computation times may further decrease in the future with the availability of new computing hardware.  However, the relative computational complexity and hence, the relative computation time, should be independent of the hardware's computational power at equal levels of parallelization.

Testing of the best performing algorithm identified through the simulations was carried out using our F-180 robots as final validation step.  The best filter was implemented using Visual C++ and the Bayesian Filtering Library$^{\cite{GAD01_BFL}}$.  Different predefined elliptical and rectilinear trajectories were generated by the planning and control system for the practical experiment.  Non-Gaussian noise was deliberately generated by placing in the court a decoy robot with similar color patches moving along a trajectory that intersects the path of robot being tracked.  Even if the likelihood of an erroneous identification is small provided an adequate illumination and image acquisition rate, in actual competitions erroneous identification does occur in part due to particular positions that the fastly moving robots may assume with respect to the field lines, the ball or other robots.  Therefore, in order to test the filter under an extreme situation, it is desirable to force the occurrence of non-Gaussian noise using a decoy robot to {\em distract} the tracker.  The next section presents and discusses the simulation and practical results obtained.

\begin{figure}[H]
\begin{center}
\small\begin{tabular}{ccc}
\rule[0mm]{0mm}{0mm}\parbox[t]{0.5\textwidth}{\hspace{-1cm}\vspace*{0cm}
\includegraphics[scale=0.65]{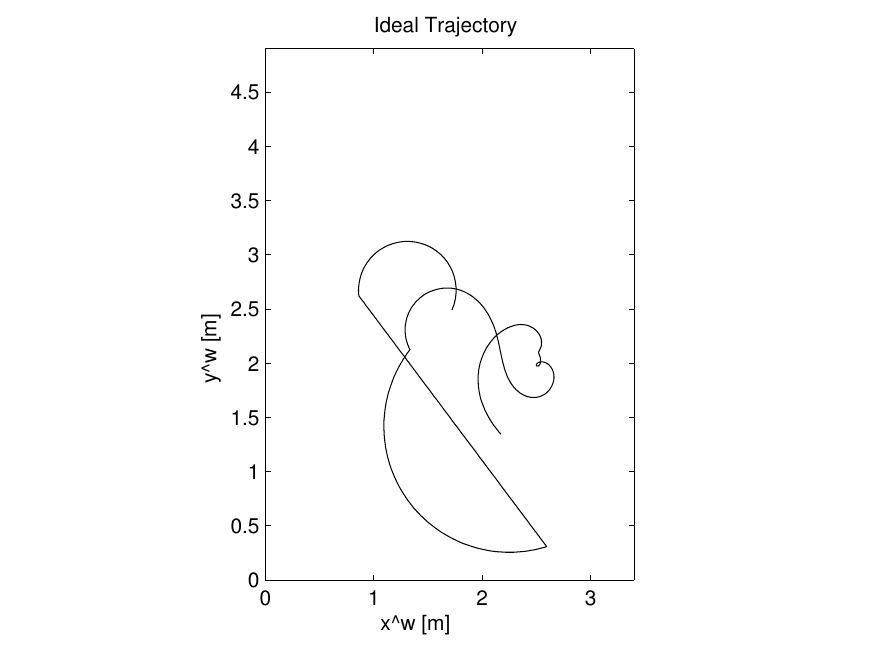}}
& \rule[0mm]{0mm}{0mm}\parbox[t]{0.5\textwidth}{\hspace{-1.5cm}\vspace*{0cm}\includegraphics[scale=0.65]{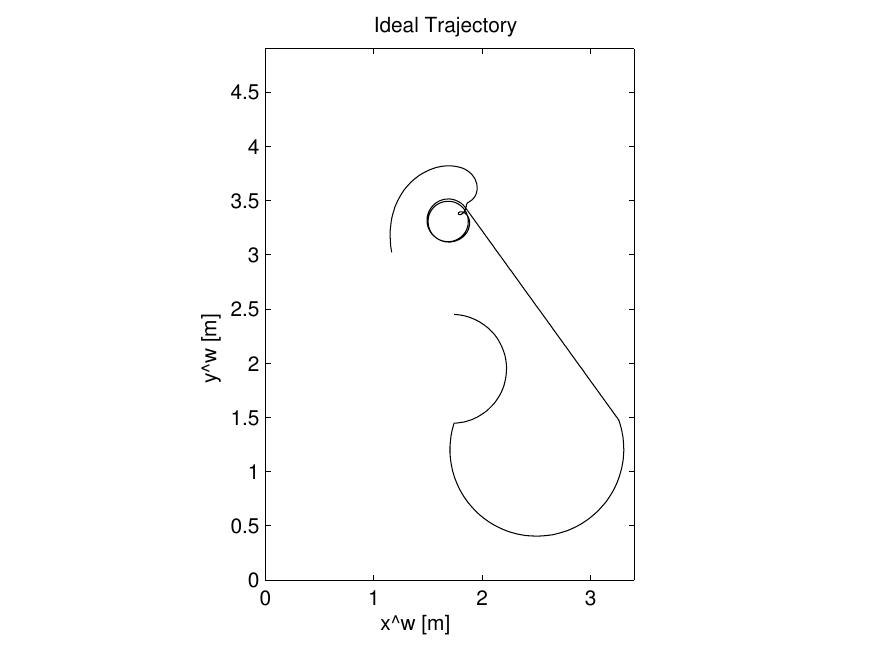}} \\
\rule[0mm]{0mm}{0mm}\parbox[t]{0.5\textwidth}{\hspace{-1cm}\vspace*{0cm}\includegraphics[scale=0.65]{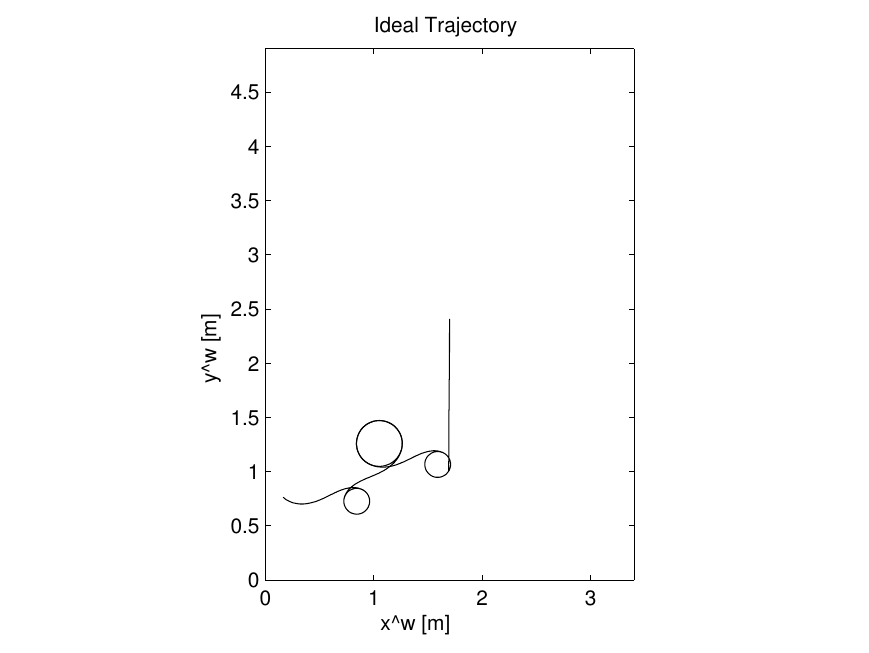}}
&\rule[0mm]{0mm}{0mm}\parbox[t]{0.5\textwidth}{\hspace{-1.5cm}\vspace*{0cm}\includegraphics[scale=0.65]{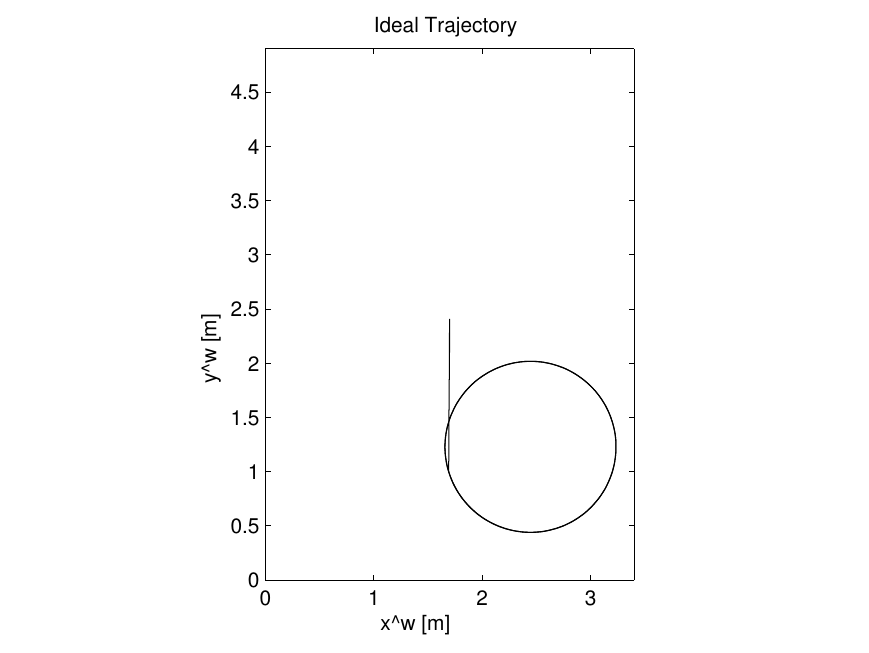}} \\
\rule[0mm]{0mm}{0mm}\parbox[t]{0.5\textwidth}{\hspace{-1cm}\vspace*{0cm}\includegraphics[scale=0.65]{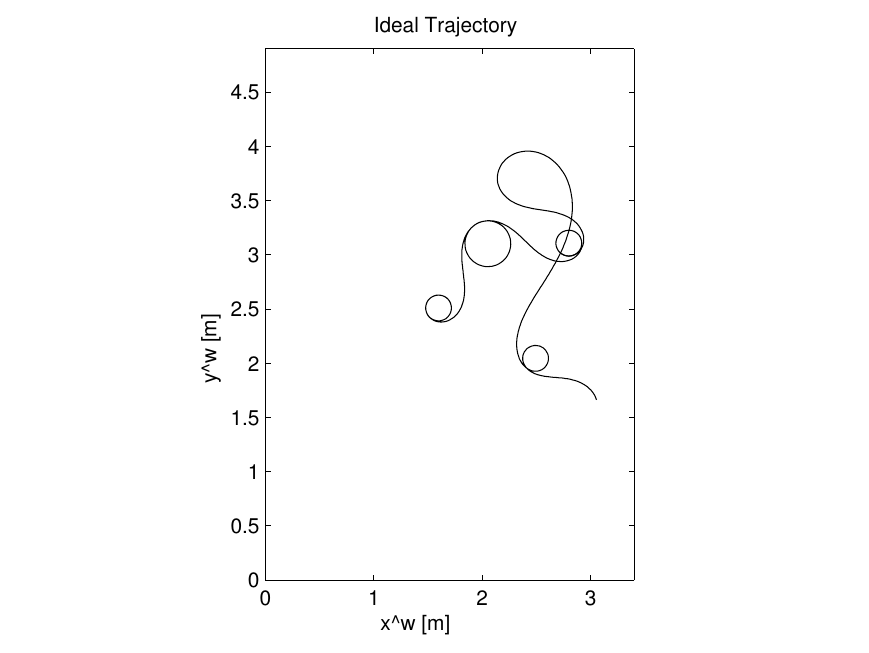}}
&\rule[0mm]{0mm}{0mm}\parbox[t]{0.5\textwidth}{\hspace{-1.50cm}\vspace*{0cm}\includegraphics[scale=0.65]{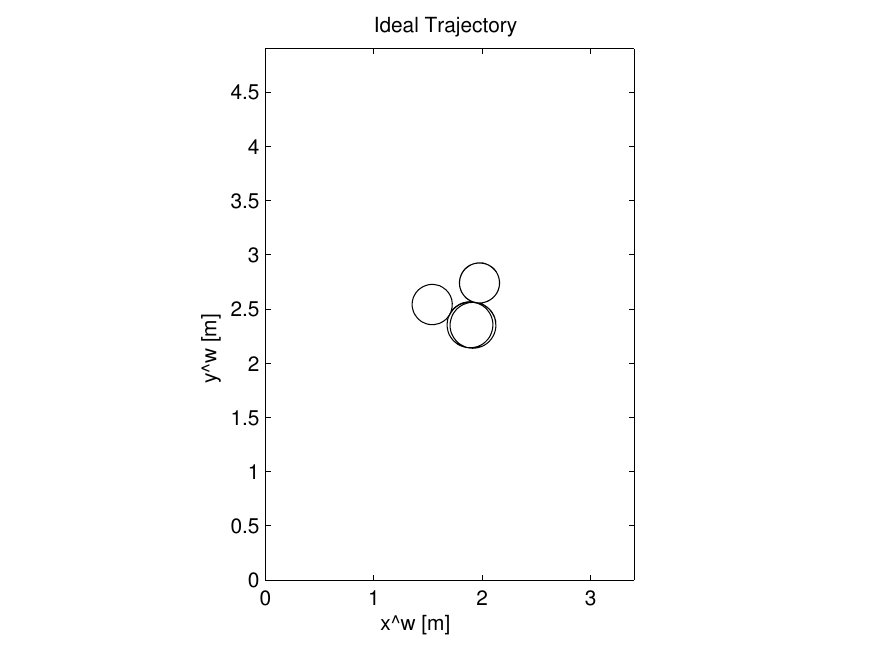}} \\
\end{tabular}
\end{center}
\vspace*{-1.5em}
\caption{Six different reference trajectories employed in the simulations.}
\label{fig:diff_traj}
\end{figure}

\begin{figure}[htb]
\begin{center}
\begin{tabular}{cc}
\rule[0mm]{0mm}{0mm}\parbox[t]{0.5\textwidth}{\hspace{-1cm}\vspace*{0cm}
\includegraphics[scale=0.65]{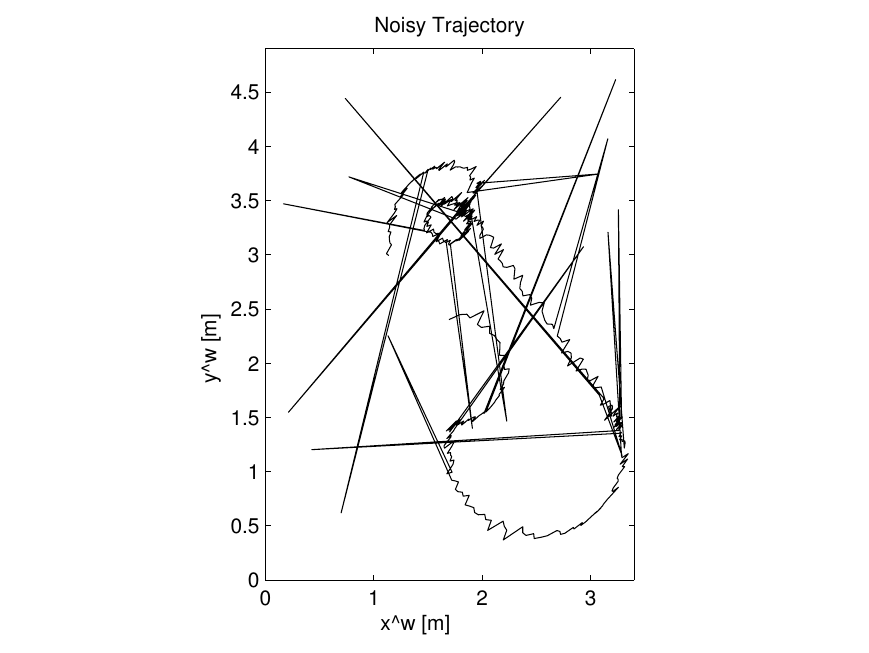}}
&\rule[0mm]{0mm}{0mm}\parbox[t]{0.5\textwidth}{\hspace{-1.5cm}\vspace*{0cm}\includegraphics[scale=0.65]{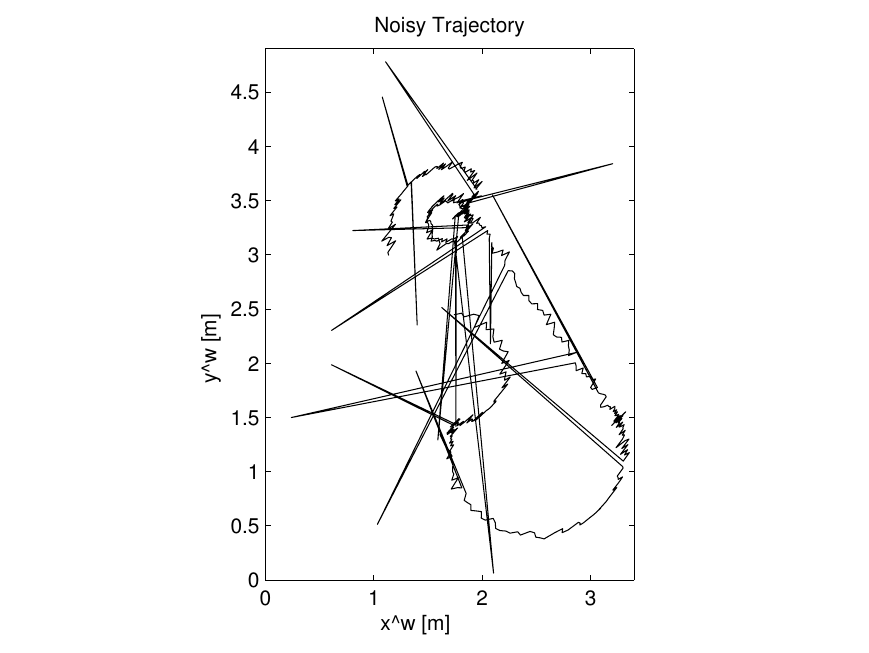}} \\
\rule[0mm]{0mm}{0mm}\parbox[t]{0.5\textwidth}{\hspace{-1cm}\vspace*{0cm}\includegraphics[scale=0.65]{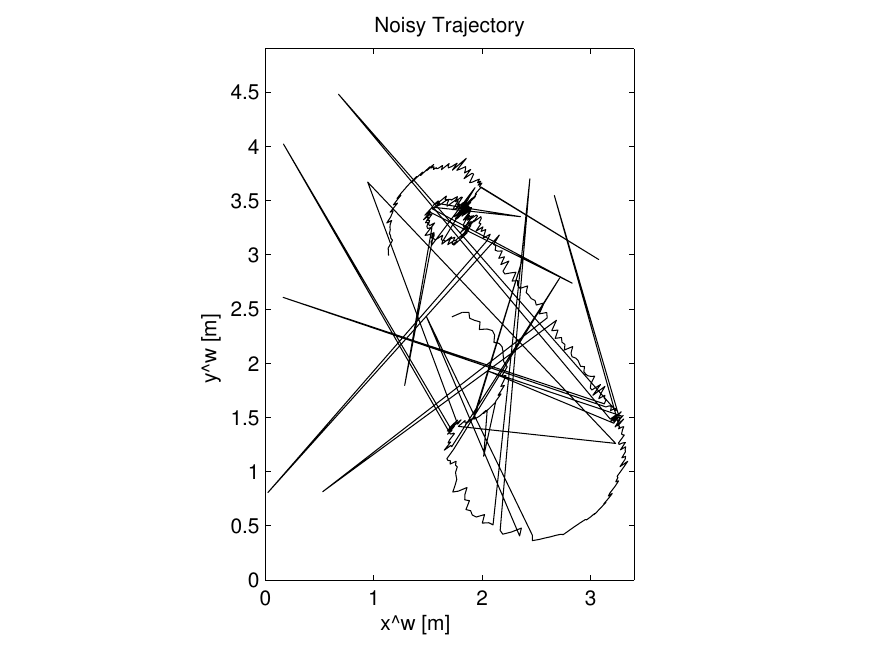}}
&\rule[0mm]{0mm}{0mm}\parbox[t]{0.5\textwidth}{\hspace{-1.5cm}\vspace*{0cm}\includegraphics[scale=0.65]{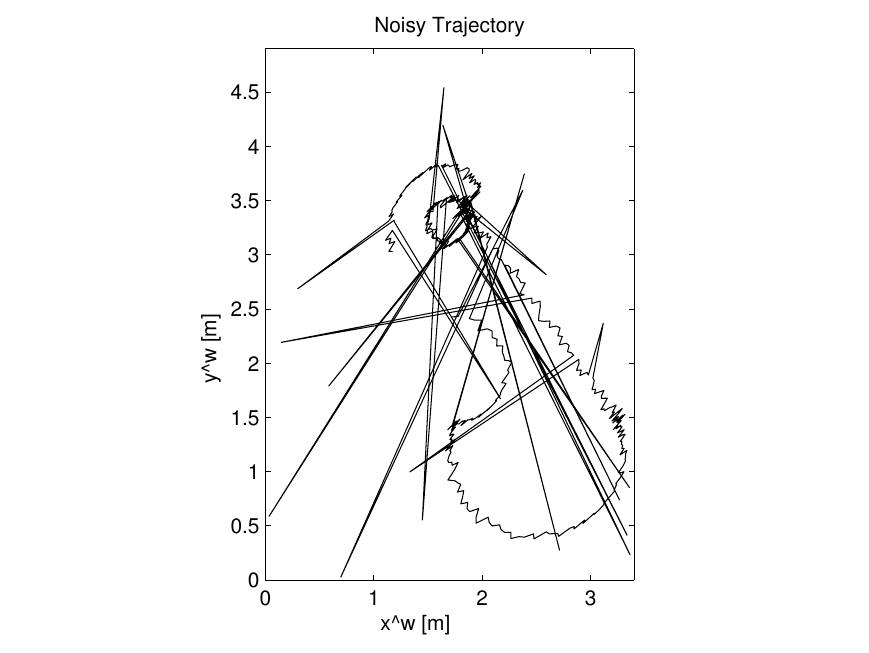}} \\
\end{tabular}
\end{center}
\vspace*{-1.5em}
\caption{Examples of the measured noisy trajectory corresponding to trajectory in the first row, second column of fig.~\ref{fig:diff_traj} for different noise realizations.}
\label{fig:Ntrj}
\end{figure}

\section{Experimental Results}\label{sec:results}
The average position and heading RMSE obtained with the different filters for 1, 4 and 8 frame prediction horizons are presented in tables~\ref{tab:result1},~\ref{tab:result2} and ~\ref{tab:result3}, respectively.  The 1, 4 and 8 frame periods correspond to 0.033, 0.133, 0.267 seconds, respectively, since the sampling period $T=1/30$ seconds is defined by the standard frame rate of the video camera.  Tables~\ref{tab:result1},~\ref{tab:result2} and~\ref{tab:result3} include RMSE values for the different filters tested with and without the Improbability Filter (IF).  As earlier mentioned, the IF is not employed with the Particle Filter (PF), because the PF does not require the Gaussian noise assumption and hence it should be able to reject non-Gaussian disturbances without necessitating the pre-filtering stage in the form of the IF.  The PF was implemented with 600 particles.

The relative computation time $T_c/{T_c}_{min}$ in tables~\ref{tab:result1},~\ref{tab:result2} and~\ref{tab:result3} is expressed relative to the fastest method, i.e. the computation time $T_c$ required by each method is divided by the computation time of the fastest method ${T_c}_{min}$.  The latter is a reasonable quantification of the relative computational complexity of the different filters implemented.  For example, table~\ref{tab:result2} shows that the 4-frame ahead trajectory predictions using the PF took on average approximately 20.4 times longer than the same 4-frame ahead predictions using the EKF with the IF.  As reference, the EKF 4-frame ahead prediction of the total 601 points of the robot's trajectory took on average 789.84 ms running in Matlab 7.0 under Windows XP on a computer with an AMD Turion 64 bits processor and 1 GB RAM.  This means that each 4-frames ahead prediction using the EKF takes approximately 1.31 ms ($= 789.84/601\,[ms/cycle]$).  Since this value is smaller than the 33.33 ms acquisition period, the EKF is a suitable candidate for real-time implementation.

The numerical simulations also revealed that there is an extra 8.9\% computational cost in the 1-frame ahead prediction when the EKF is used together with the IF.  This extra computational cost decreases to 6.3\% and 5.3\% for the 4 and 8-frame prediction horizons, respectively.  For the most computationally demanding filter, the SPPF, the additional computational cost introduced by the IF is negligible and around 2.7\% for the three prediction horizons.  This is expectable because in the case of the SPPF its complexity makes its cost quite considerable with respect to the small cost of the simple IF.

Tables~\ref{tab:result1} and~\ref{tab:result2} show that the smallest position RMSE for the 1 and 4-frame prediction is obtained with the PF.  For the 8-frame ahead prediction, the results in table~\ref{tab:result3} indicate that the SPPF with IF yields slightly better results than the PF, however the SPPF with IF is 7.1 times computationally more demanding than the PF, and hence, the PF is still a better choice.  Other filters that exhibit good accuracies in the position prediction were the GMSPPF and SPPF.  However, these do not perform as well as the CDKF, SR-UKF or SR-CDKF for angle predictions 4 or 8-frames ahead. On the other hand, it is to be noted that the computational demand of the PF is about five times greater per prediction cycle than that of the simpler EKF with IF.  The latter offers the best trade-off between accuracy and speed.  This is  because the EKF with IF was ranked second or third in the 1, 4 or 8-frame prediction in terms of the position RMSE, yet it was ranked first in terms of computational speed.

{
\renewcommand{\baselinestretch}{1.1}
\begin{table}[H]
\caption{Results for the 1-frame ahead prediction.}
\label{tab:result1}
\vspace*{-.8\baselineskip}
\begin{center} 
{\small
\begin{tabular}{lcccccc}
\hline
& \multicolumn{2}{c}{RMSE w/IF} & \multicolumn{2}{c}{RMSE wo/IF} & 
 $T_c/{T_c}_{min}$ & $T_c/{T_c}_{min}$\\
\textsc{Filter} & {\em Position} [$m$] & {\em Heading} [$rad$]& {\em Position} [$m$] & {\em Heading} [$rad$] & w/IF & wo/IF\\ \hline
EKF     & 0.0048$^b$ & 0.2326 & 0.0229$^b$ & 0.2674 & 1.000  & 1.000\\
DD1     & 0.0434 & 0.1386 & 0.0491$^c$ & 0.1951 & 3.566  & 3.729\\
DD2     & 0.0342 & 0.1363 & 0.0491 & 0.1947 & 3.566  & 3.729\\
UKF     & 0.0379 & 0.0398$^c$ & 0.1101 & 0.1399$^c$ & 3.758  & 3.971\\
CDKF    & 0.0105$^c$ & 0.0518 & 0.1100 & 0.1398$^b$ & 3.727  & 3.960\\
SR-UKF  & 0.0397 & 0.0364$^a$ & 0.1099 & 0.1399$^c$ & 3.872  & 3.956\\
SR-CDKF & 0.1656 & 0.0941 & 0.1099 & 0.1402 & 3.734  & 3.956\\
PF      &  ---   &  ---   & 0.0014$^a$ & 0.0327$^a$ &  ---   & 4.901\\
GMSPPF  & 0.0034$^a$ & 0.0392$^b$ & 0.0637 & 1.0703 & 31.973 & 33.855\\
SPPF    & 0.0690 & 0.1042 & 0.1103 & 0.1404 & 42.687 & 45.996\\
\hline
\multicolumn{7}{l}{$a=$ best, $b=$ second best, $c=$ third best result.}
\end{tabular}
}
\end{center}
\end{table}

\begin{table}[H]
\caption{Results for the 4-frame ahead prediction.}
\label{tab:result2}
\vspace*{-.8\baselineskip}
\begin{center} 
{\small
\begin{tabular}{lcccccc}
\hline
& \multicolumn{2}{c}{RMSE w/IF} & \multicolumn{2}{c}{RMSE wo/IF} & 
 $T_c/{T_c}_{min}$ & $T_c/{T_c}_{min}$\\
\textsc{Filter} & {\em Position} [$m$] & {\em Heading} [$rad$]& {\em Position} [$m$] & {\em Heading} [$rad$] & w/IF & wo/IF\\ \hline
EKF     & 0.0234$^b$ & 0.2078 & 0.0478$^b$ & 0.2510 & 1.000   & 1.000 \\
DD1     & 0.0548 & 0.1504 & 0.1232 & 0.1351 & 2.499   & 2.566\\
DD2     & 0.0658 & 0.1535 & 0.1232 & 0.1360$^b$ & 1.749   & 2.751\\
UKF     & 0.1187 & 0.0541 & 0.0617$^c$ & 0.2108 & 3.790   & 3.965\\
CDKF    & 0.0561 & 0.0325$^a$ & 0.0617 & 0.2115 & 3.594   & 3.756\\
SR-UKF  & 0.0515$^c$ & 0.0363$^c$ & 0.1232 & 0.1356 & 3.582   & 3.736\\
SR-CDKF & 0.0618 & 0.0333$^b$ & 0.1231 & 0.1361$^c$ & 3.582   & 3.736\\
PF      &  ---   &  ---   & 0.0191$^a$ & 0.1189$^a$ &  ---    & 20.424\\
GMSPPF  & 0.0927 & 0.1353 & 0.1157 & 0.2110 & 112.630 & 116.457\\
SPPF    & 0.0192$^a$ & 0.1483 & 0.1160 & 0.2137 & 145.480 & 157.970\\
\hline
\multicolumn{7}{l}{$a=$ best, $b=$ second best, $c=$ third best result.}
\end{tabular}
}
\end{center}
\end{table}

\begin{table}[H]
\caption{Results for the 8-frame ahead prediction.}
\label{tab:result3}
\vspace*{-.8\baselineskip}
\begin{center} 
{\small
\begin{tabular}{lcccccc}
\hline
& \multicolumn{2}{c}{RMSE w/IF} & \multicolumn{2}{c}{RMSE wo/IF} & 
 $T_c/{T_c}_{min}$ & $T_c/{T_c}_{min}$\\ 
\textsc{Filter} & {\em Position} [$m$] & {\em Heading} [$rad$]& {\em Position} [$m$] & {\em Heading} [$rad$] & w/IF & wo/IF\\ \hline
EKF     & 0.0611$^c$ & 0.2138 & 0.0828$^b$ & 0.2577 & 1.000   & 1.000\\
DD1     & 0.0884 & 0.1541 & 0.0928 & 0.2145 & 2.220   & 2.261\\
DD2     & 0.0582$^b$ & 0.1373 & 0.0927$^c$ & 0.2143 & 2.210   & 2.261\\
UKF     & 0.0812 & 0.0784 & 0.1549 & 0.1402$^c$ & 3.874   & 4.024\\
CDKF    & 0.1208 & 0.0400$^a$ & 0.1549 & 0.1395$^b$ & 3.561   & 3.696\\
SR-UKF  & 0.1151 & 0.0438$^b$ & 0.1548 & 0.1389$^a$ & 3.569   & 3.670\\
SR-CDKF & 0.1468 & 0.0704$^c$ & 0.1549 & 0.1404 & 3.569   & 3.670\\
PF      &  ---   &  ---   & 0.0416$^a$ & 0.3107 &  ---    & 28.757\\
GMSPPF  & 0.2269 & 0.4195 & 0.1316 & 0.4340 & 159.053 & 162.990\\
SPPF    & 0.0308$^a$ & 0.3835 & 0.1318 & 0.4364 & 204.304 & 221.425\\
\hline
\multicolumn{7}{l}{$a=$ best, $b=$ second best, $c=$ third best result.}
\end{tabular}
}
\end{center}
\end{table}
} 

Figure~\ref{fig:cln_trj} shows one of the six noise-free reference trajectories (see fig.~\ref{fig:diff_traj}) employed in the simulations, while fig.~\ref{fig:noisy_trj} shows a noisy measurement corresponding to the trajectory in fig.~\ref{fig:cln_trj}. The 4-frame ahead predictions resulting from the application of the EKF without the IF and with the IF are shown in figures~\ref{fig:EKF_woIF} and~\ref{fig:EKF_wIF}, respectively.  It is possible to clearly see from the comparison of figures~\ref{fig:EKF_woIF} and~\ref{fig:EKF_wIF} that the prediction using the EKF with the IF to reject non-Gaussian noise is substantially better, while disastrous results are obtained with the EKF alone.  In terms of the position and heading RMSE values presented in table~\ref{tab:result2}, the PF is slightly better than the EKF with IF, however the PF is not capable of generating reasonably smoothed trajectories as can be appreciated from the comparison of figures~\ref{fig:EKF_woIF} and~\ref{fig:PF_4fr}.

\begin{figure}[htbp]
\begin{center}
\includegraphics[scale=0.75]{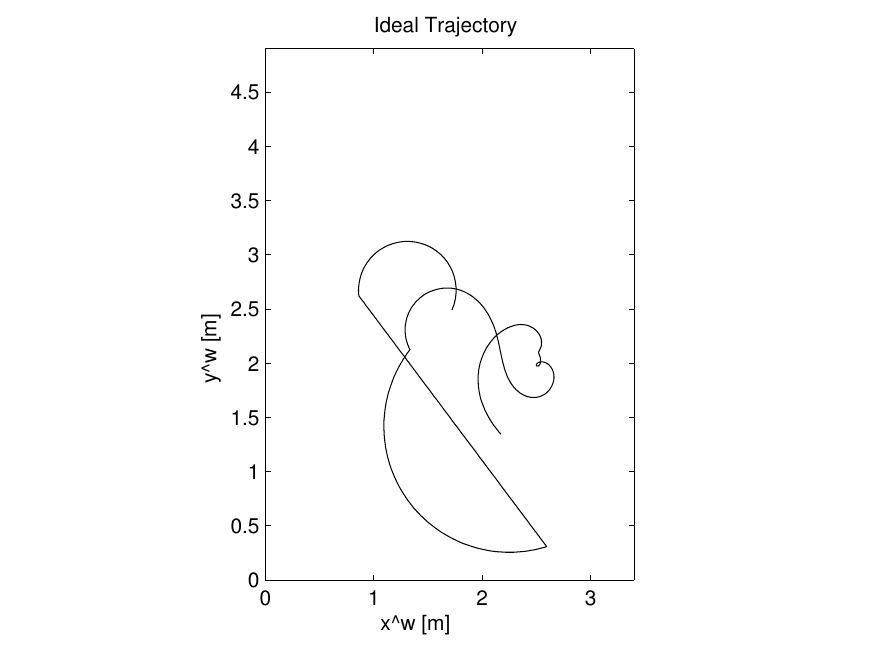}
\end{center}
\vspace*{-1.5em}
\caption{Noise-free reference trajectory.}
\label{fig:cln_trj}
\end{figure}

\begin{figure}[htbp]
\begin{center}
\includegraphics[scale=0.75]{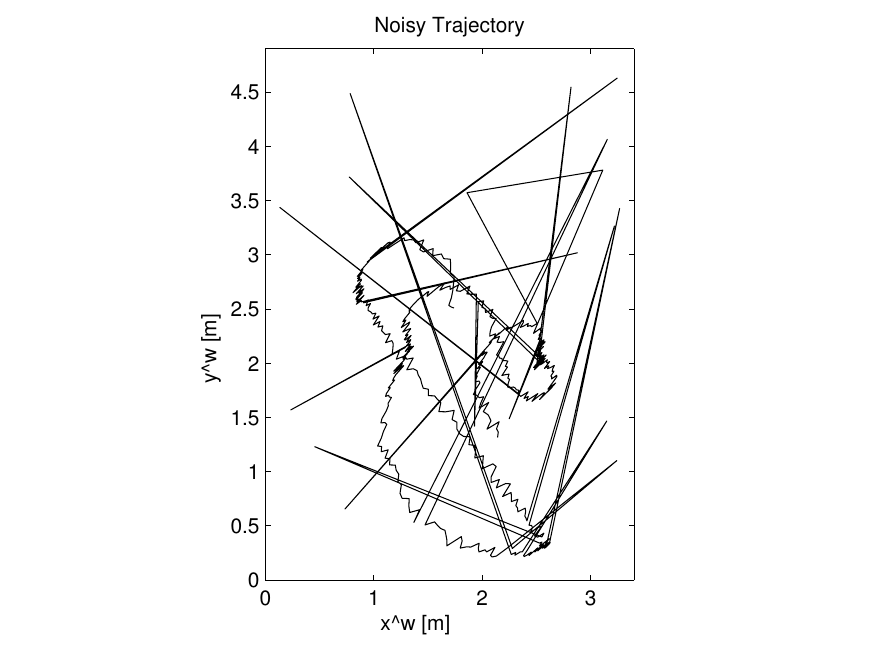}
\end{center}
\vspace*{-1.5em}
\caption{Measured trajectory.}
\label{fig:noisy_trj}
\end{figure}

\begin{figure}[htbp]
\begin{center}
\includegraphics[scale=0.75]{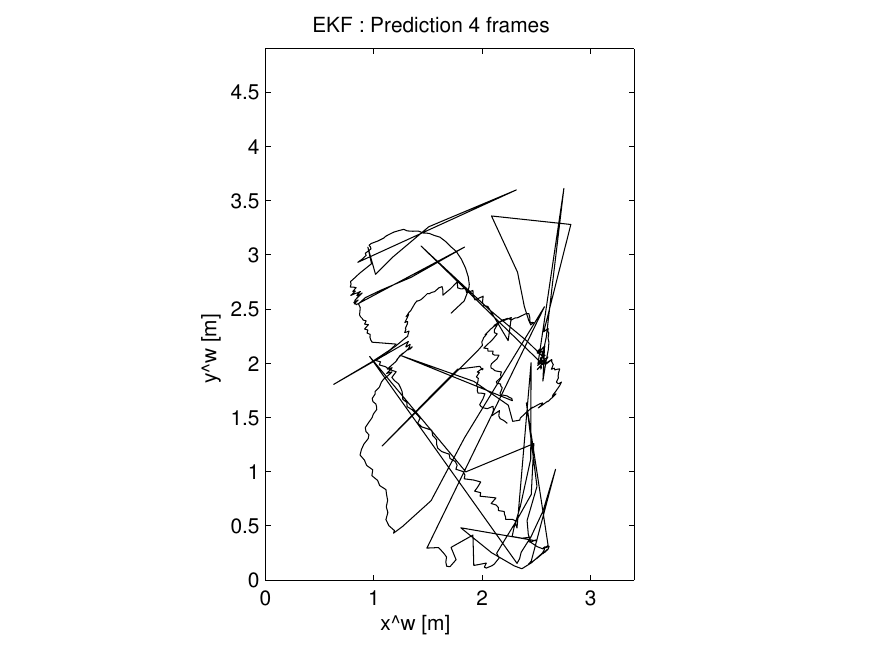}
\end{center}
\vspace*{-1.5em}
\caption{Predicted trajectory 4-frames ahead using the EKF without the IF.}
\label{fig:EKF_woIF}
\end{figure}

\begin{figure}[htbp]
\begin{center}
\includegraphics[scale=0.75]{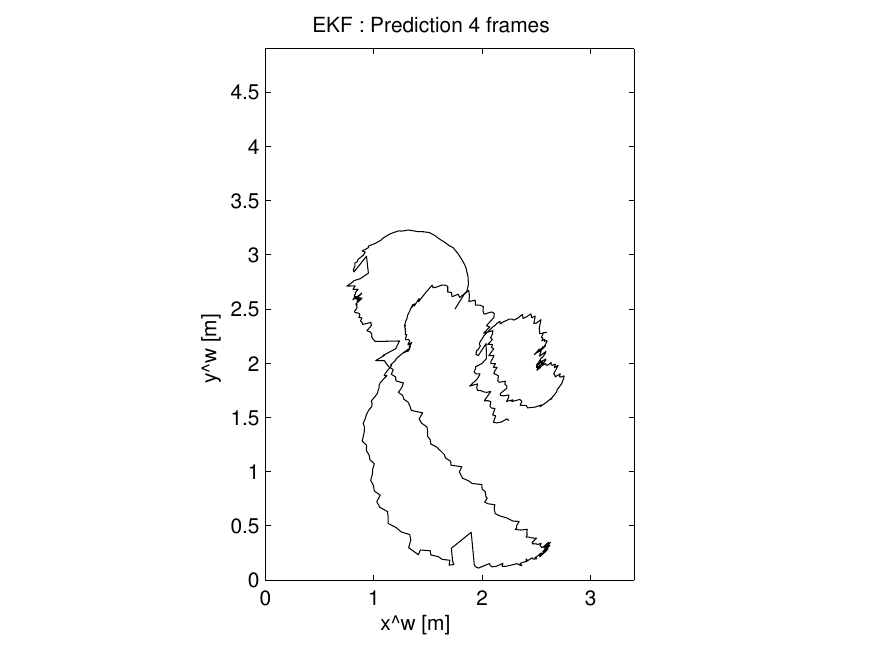}
\end{center}
\vspace*{-1.5em}
\caption{Predicted trajectory 4-frames ahead using the EKF with the IF.}
\label{fig:EKF_wIF}
\end{figure}

\begin{figure}[htbp]
\begin{center}
\includegraphics[scale=0.75]{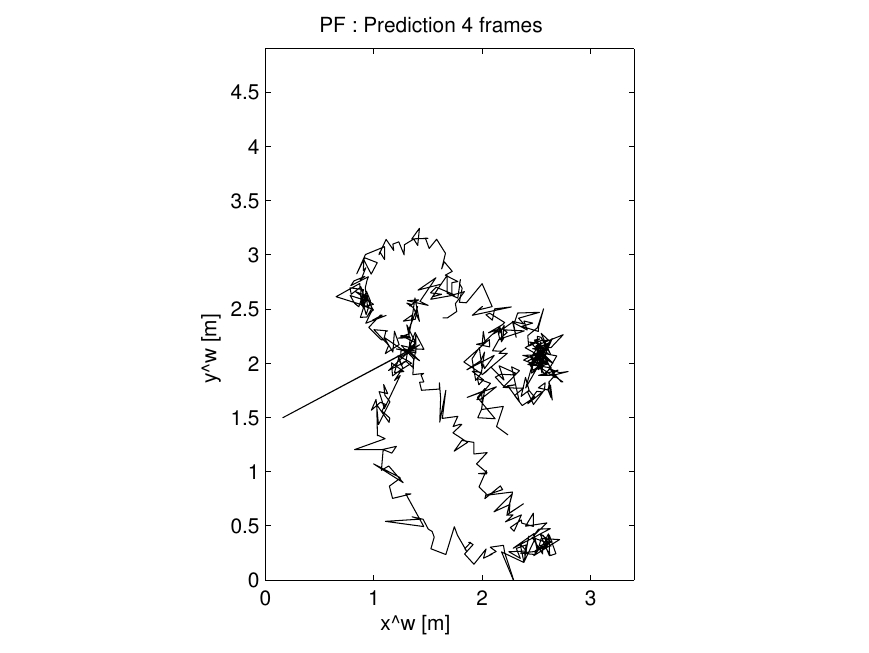}
\end{center}
\vspace*{-1.5em}
\caption{Predicted trajectory 4-frames ahead using the PF.}
\label{fig:PF_4fr}
\end{figure}

The EKF with the IF was chosen for validation using the robots in a real court because the EKF demonstrated a reasonable accuracy and smallest computational cost.  Figure~\ref{fig:Cpp1} shows the results of one of the tests in which the robot was programmed to follow a simple straight line parallel to the $\mathbf{y^w}$ axis back and forth.  The motion was repeated four times and lasted approximately 11 seconds.  From fig.~\ref{fig:Cpp1} it is possible to observe that the filter is capable of overcoming the non-Gaussian noise and track the programmed reference trajectory.  It is to be noted that the zigzagging of the estimation about the reference trajectory is mainly due to the fact that the robot used for testing had low cost gearboxes and wheels, which generated a jerky motion difficult to control.
\begin{figure}[htbp]
\begin{center}
\hspace*{-1.8cm}\includegraphics[scale=0.52]{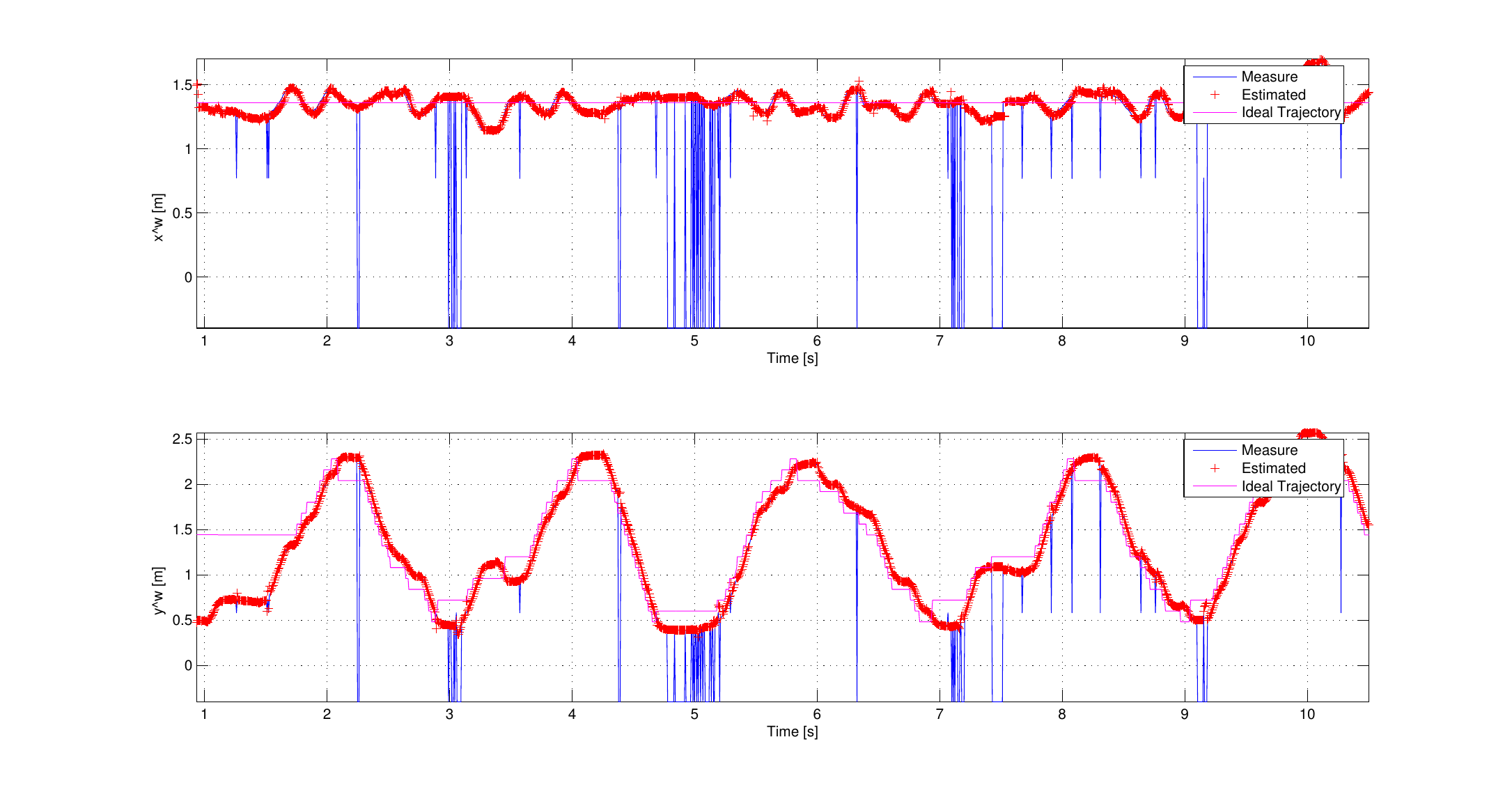}
\end{center}
\vspace*{-1.5em}
\caption{Trajectory components $x^w$ and $y^w$ of a real F-180 robot predicted 4-frames ahead using the EKF with the IF.}
\label{fig:Cpp1}
\end{figure}

\section{Conclusions and Future Work}\label{sec:conclusions}
The performance of ten different Bayesian filters was evaluated in terms of their ability to handle non-Gaussian noise and to make accurate state predictions.  This work's contribution is in the identification of the best filtering techniques capable of handling non-Gaussian disturbances or noise with a comparatively low computational effort in order to make accurate state predictions.  The ability to make a good prediction of a system's state, rather than to simply obtain a good state estimate at the current time, is fundamental in many autonomous systems applications that involve expensive computations to perform perception, control and decision tasks.  Such expensive computations introduce delays that limit response times of the closed-loop system.  Therefore, accurate state predictors become essential to cope with the delays.

The filtering techniques evaluated have been proposed during the last decades, however no exhaustive comparison was found in the literature.  Some of the techniques belong to the family of the Kalman filter and its modifications, while other to the family of Sequential Monte Carlo sampling schemes.  A third group is composed by hybrid filters, which combine the Kalman filtering approaches with Sequential Monte Carlo techniques.

The results obtained indicate that in general both the simple EKF with the IF and the PF yield accurate predictions. However, the PF requires a significantly larger computational effort.  This larger computational cost could nevertheless be decreased by sacrificing accuracy if the number of particles in the PF implementation is reduced.   The PF implemented in the simulations used 600 particles and had a slightly better accuracy than that of the EKF with the IF, however the PF was about 5 times slower than the EKF with IF per sampling period.  This means that for prediction horizons of $n$ sampling periods, the PF would take $5n$ times longer than the simpler EKF with IF.  Some other methods evaluated also satisfy the accuracy requirements of the application presented, but their computational complexity and comparatively smaller accuracy makes them a less reasonable choice.  In particular, the GMSPPF and SPPF are computationally expensive and less suitable for applications requiring online filtering.  Future availability of more powerful computers may allow to reduce the computation time required by these methods, however their advantages in the context of tracking multiple robots is not clear.

It is also worth noting that the inclusion of the Improbability Filter allows for significant prediction improvements in all the different filters implemented that assume Gaussian noises (all filters except for the Particle Filter).  A minor drawback of the Improbability Filter is that it requires the appropriate threshold value to be found empirically.  On the other hand, the right tuning of the hybrid SPKF combined with SMC sampling methods may be very difficult because a slight modification of the filter parameters may result in non positive-definite state covariance matrices.  This is because the sigma points are deterministically calculated in terms of the square root of the weighted state covariance matrix$^{\cite{MER04A}}$.  The approaches preserve the first two statistical moments of a not necessarily Gaussian distribution and ensure that the predicted covariance stays semi-positive definite, unlike previous methods which could yield negative definite predicted covariances.  However, the choice of the filter parameters and the numerical round-off errors that accumulate throughout the simulation may render the predicted covariance matrix to be not strictly positive definite.  In such a case, it may not be possible to compute the matrix square root, even if the numerically efficient Cholesky factorization$^{\cite{PRE92}}$ is employed.  To overcome this problem, the state covariance matrix was reset whenever it became non-positive definite.  However, this action also limits the performance of the hybrid approaches.  

Future work considers the implementation of multiple hypothesis tracking methods to increase the robustness of the planning and control system that drives the team of robots, the development of techniques for automatically adjusting the IF parameters so as to maximize the rejection of false positives, and the comparison of the Bayesian methods with non-Bayesian techniques, such as approaches based on neural networks.  

\section*{Acknowledgements}
This project has been supported by the National Commission for Science and Technology Research of Chile (Conicyt) under Fondecyt Grant 11060251.

\section*{References}
\begin{enumerate}
\item \label{THR05} 
S. Thrun, W. Burgard, D. Fox, {\em Probabilistic Robotics} (MIT Press, Cambridge, MA, 2005).
\item \label{RIS04} 
B. Ristic, S. Arulampalam and N. Gordon, {\em Beyond the Kalman Filter: Particle Filters for Tracking Applications} (Artech House, Norwood, MA, 2004).

\item \label{SSE04} Special Issue on Sequential State Estimation, {\em Proceedings of the IEEE} {\bf 92}(3), (2004). 

\item \label{TSA03} 
M.-C. Tsai, K.-Y. Chen, M.-Y. Cheng and K. C. Lin, ``Implementation of a real-time moving object tracking system using visual servoing,'' {\em Robotica} {\bf 21}(6), 615--625 (2003). 
\item \label{HUE02} 
C. Hue, J.P. Le-Cadre and P. Perez, ``Sequential Monte Carlo methods for multiple target tracking and data fusion,'' {\em IEEE Trans. Signal Processing} {\bf 50}(2), 309--325 (2002).
\item \label{ORT02} 
M. Orton and W. Fitzgerald, ``A Bayesian approach to tracking multiple targets using sensor arrays and particle filters,'' {\em IEEE Trans. Signal Processing} {\bf 50}(2) 216--223 (2002). 

\item \label{DOU00} A. Doucet, S. Godsill and C. Andrieu, ``On sequential Monte Carlo sampling methods for Bayesian filtering,'' {\em Statistics and Computing} {\bf 10}(3), 197--208 (2000).

\item \label{BAR98} 
Y. Bar-Shalom and X. Rong Li, {\em Estimation and Tracking: Principles, Techniques, and Software} (Ybs Publishing, 1998).

\item \label{ASK00} 
F. Askary, N. T. Sullivan, ``Importance of measurement accuracy in statistical pprocess control,'' {\em Proc. SPIE Metrology, Inspection, and Process Control for Microlithography XIV} (N. T. Sullivan ed.) {\bf 3998}, (2000) 546--554. 
\item \label{MER04A} 
R. van der Merwe, E. A. Wan and S. Julier, ``Sigma-Point Kalman Filters for Nonlinear Estimation and Sensor-Fusion: Applications to Integrated Navigation'', {\em Proceedings of the AIAA Guidance, Navigation and Control Conference}, (2004) 5120.

\item \label{GLO03} 
A. Gloye, M. Simon, A. Egorova, F. Wiesel, O. Tenchio, M. Schreiber, S. Behnke and R. Rojas, {\em Predicting away robot control latency}, (Technical Report B-08-03, Freie Universit\"at Berlin,  Fachbereich Mathematik und Informatik, 2003). 
\item \label{BOU01} 
D. Bouvet and G. Garcia, ``GPS latency identification by Kalman filtering,'' {\em Robotica} {\bf 18}(5), 475--485 (2001).
\item \label{DON07}
F. Donoso-Aguirre, J.-P. Bustos-Salas, M. Torres-Torriti and A. Guesalaga, ``Mobile robot localization using the Hausdorff distance,'' {\em To appear in Robotica, published online by Cambridge University Press, 12 Jul 2007}. 

\item \label{AND05} 
B.D.O. Anderson and J.B. Moore, {\em Optimal Filtering}, (Dover Publications, Mineola, NY, 2005).

\item \label{KAL60} 
R.E. Kalman, ``New approach to linear filtering and prediction problems,'' {\em Trans. of the ASME -- Journal of Basic Engineering}, {\bf 82}, 34--45 (1960).

\item \label{JAZ70} 
A. H. Jazwinski, {\em Stochastic Processes and Filtering Theory}, (Academic Press, 1970).
\item \label{JUL97} 
S. J. Julier and J.K. Uhlmann, ``A new extension of the Kalman filter to nonlinear systems,'' {\em Proceedings of AeroSense: The 11th International Symposium on Aerospace/Defence Sensing, Simulation and Controls}, Orlando, Florida, (1997).
\item \label{ITO00} 
K. Ito and K. Xiong, ``Gaussian filters for nonlinear filtering problems,''  {\em Proceedings of the IEEE International Conference on Robotics and Automation} {\bf 45}(5), (2000) 910--927.
\item \label{NOR00} 
M. N{\o}rgaard, N.K. Poulsen and O. Ravn, ``New developments in state estimation for nonlinear systems,'' {\em Automatica} {\bf 36}(11), 1627--1638 (2000).
\item \label{MER01} 
R. Van-der-Merwe and E. Wan, ``The Square-Root Unscented Kalman Filter for state and parameter estimation,'' {\em Proceedings of the IEEE International Conference on Acoustics, Speech, and Signal Processing} {\bf 6}, (2001) 3461--3464.
\item \label{MER00} 
R. Van-der-Merwe, A. Doucet, N. de Freitas and E. Wan, ``The Unscented Particle Filter,'' {\bf In:} {\em Advances in Neural Information Processing Systems} (T. K. Leen, T. G. Dietterich and V. Tresp eds.), 584--590 (MIT Press, Cambridge, MA, 2000).

\item \label{MER03} 
R. Van-der-Merwe and E. Wan, ``Gaussian mixture Sigma-Point Particle Filters for sequential probabilistic inference in dynamic state-space models,'' {\em Proceedings of the International Conference on Acoustic, Speech and Signal Processing} (2003) 701--704.

\item \label{KOT01} 
J.H. Kotecha and P.M. Djuric, ``Gaussian sum Particle Filtering for dynamic state space models,'' {\em Proceedings of IEEE International Conference on Acoustics, Speech, and Signal Processing 2001} (2001) 3465--3468. 

\item \label{CUI05} 
N. Cui, L. Hong, J. R. Layne, ``A comparison of nonlinear filtering approaches with an application to ground target tracking,'' {\em Signal Processing, Elsevier}, {\bf 85}(8), 1469--1492 (2005). 

\item \label{WRI03} 
R. Wright, S. R. Maskell, M. Briers, ``Comparison of Kalman-based methods with Particle Filters for raid tracking'', {\em Practical Bayesian Statistics 5}, (2003). 
\item \label{YUE02} 
D. C. K. Yuen, B. A. MacDonald, ``A comparison between Extended Kalman Filtering and Sequential Monte Carlo techniques for simultaneous localisation and map-building,'' {\em Proceedings of the 2002 Australasian Conference on Robotics and Automation} (2002) 111--116. 

\item \label{CUE05A} 
E. V. Cuevas, D. Zald\'ivar, and R. Rojas, {\em Kalman filter for vision tracking}, (Technical Report B-05-12, Freie Universit\"at Berlin, Fachbereich Mathematik und Informatik, 2005).
\item \label{CUE05B} 
E. V. Cuevas, D. Zald\'ivar, and R. Rojas, {\em Particle filter for vision tracking}, (Technical Report B-05-13, Freie Universit\"at Berlin, Fachbereich Mathematik und Informatik, 2005).

\item \label{MET49} 
N. Metropolis and S. Ulam, ``The Monte Carlo method,'' {\em Journal of the American Statistical Association}, {\bf 44}(247), 335--341 (1949).
\item \label{GOR93} 
N. J. Gordon, D. J. Salmond and A. F. M. Smith, ``Novel approach to nonlinear/non-Gaussian Bayesian state estimation,'' {\em IEE Proc. F} {\bf 140} (2), 107--113 (1993). 

\item \label{GOR97} 
M. J. Goris, D. A. Gray, I. M. Y. Mareels, ``Reducing the computational load of a Kalman filter,'' {\em IEE Electronics Letters} {\bf 33} (18), 1539--1541 (1997). 
\item \label{KAR05} 
R. Karlsson, T. Sh{\"o}n, F. Gustafsson, ``Complexity analysis of the marginalized particle filter,'' {\em IEEE Transactions on Signal Processing} {\bf 53} (11), 4408--4411 (2005). 

\item \label{BRO02} 
B. Browning, M. Bowling and M. Veloso, ``Improbability filtering for rejecting false positive,'' {\em Proceedings of the IEEE International Conference on Robotics and Automation}, (2002) 3038--3043.
\item \label{LIU03} 
Y. Liu, X. Wu, J. Zhu and J. Lew, ``Omni-directional mobile robot controller design by trajectory linearization,'' {\em Proceedings of the 2003 American Control Conference} {\bf 4}, (2003) 3423--3428.
\item \label{ANG06} 
J. Angeles, {\em Fundamentals of Robotic Mechanical Systems: Theory, Methods, and Algorithms}, (3$^{rd}$ ed., Springer, 2006).

\item \label{MER02_REBEL}
R. van der Merwe, {\em ReBEL: Recursive Estimation Bayesian Library},
(OGI School of Science \& Engineering, Oregon Health \& Science University (OHSU), 2002-2006)\\
\texttt{http://choosh.cse.ogi.edu/rebel/}, visited on February 2006.

\item \label{NOR98_KALMTOOL} 
M. N{\o}rgaard, {\em Kalmtool: version 2},\\
\texttt{http://www.iau.dtu.dk/research/control/kalmtool.html}\\
Ole Ravn and Niels Kj{\o}lstad Poulsen, {\em KalmTool: versions 3, 4},\\
\texttt{http://server.oersted.dtu.dk/www/or/kalmtool/}, visited February 2006,
(Department of Mathematical Modelling, Department of Automation, {\O}rsted - Danmarks Tekniske Universitet, 1998-2006).

\item \label{GAD01_BFL} 
K. Gadeyne, {\em BFL: Bayesian Filtering Library } 
(Department of Mechanical Engineering, Katholieke Universiteit Leuven, Belgium, 2001-2006)\\
\texttt{http://www.orocos.org/bfl},  visited on February 2006.

\item \label{PRE92} 
W. H. Press, S. A. Teukolsky, W. T. Vetterling and E. P. Flannery, {\em Numerical Recipes in C: The Art of Scientific Computing}, (2$^nd$ ed., Cambridge University Press, MA, 1992).

\end{enumerate}

\end{document}